\pdfoutput=1
\documentclass[10pt,twocolumn,letterpaper]{article}
\usepackage[pagenumbers]{cvpr}

\usepackage{amsmath,amsfonts,bm}

\def\eqref#1{equation~\ref{#1}}

\def\1{\bm{1}}

\DeclareMathAlphabet{\mathsfit}{\encodingdefault}{\sfdefault}{m}{sl}
\SetMathAlphabet{\mathsfit}{bold}{\encodingdefault}{\sfdefault}{bx}{n}

\usepackage{xcolor}

\usepackage{url}
\usepackage{booktabs,multirow}
\usepackage{graphicx}
\usepackage{colortbl}   
\usepackage{amssymb}
\usepackage{enumitem}
\usepackage{fvextra}
\usepackage{placeins}
\usepackage{microtype}
\definecolor{cvprblue}{rgb}{0.21,0.49,0.74}
\usepackage[pagebackref,breaklinks,colorlinks,allcolors=cvprblue]{hyperref}
\def\confName{CVPR}
\def\confYear{2027}
\title{VideoRNG: Relightable Neural Gaussians from Generative Video Models}

\author{Bowen Xue$^{1,2}$, Saeed Hadadan$^{1}$, Fabrice Rousselle$^{1}$,\\
Ravi Ramamoorthi$^{3}$, Zahra Montazeri$^{2}$ \& Milo\v{s} Ha\v{s}an$^{1}$\\[3pt]
$^{1}$NVIDIA\quad $^{2}$University of Manchester\quad $^{3}$University of California, San Diego
}
\hypersetup{
  pdftitle={VideoRNG: Relightable Neural Gaussians from Generative Video Models},
  pdfauthor={Bowen Xue, Saeed Hadadan, Fabrice Rousselle, Ravi Ramamoorthi, Zahra Montazeri, Milo\v{s} Ha\v{s}an}
}

\begin{document}
\raggedbottom
\setlength{\textfloatsep}{6pt plus 2pt minus 2pt}
\setlength{\dbltextfloatsep}{6pt plus 2pt minus 2pt}

\makeatletter
\g@addto@macro\@maketitle{%
  \par\noindent
  \begin{minipage}{\textwidth}
  \centering
  \captionsetup{hypcap=false}
  \includegraphics[width=0.95\linewidth]{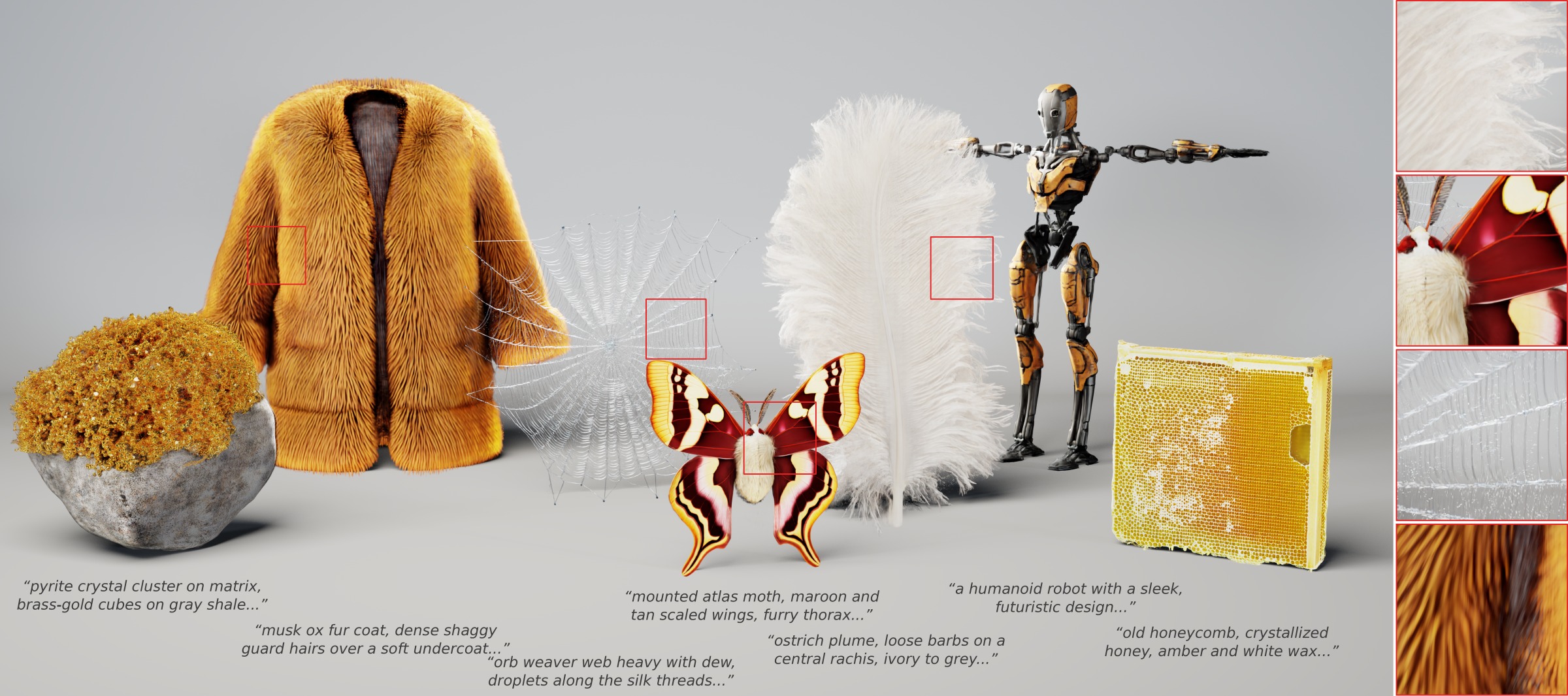}
  \captionof{figure}{VideoRNG generates relightable 3D assets from a text prompt or an image. Here we show rtRNG assets rendered together in a single scene using a full path tracer: a fur coat, pyrite crystal cluster, moth, spider web, ostrich feather, robot figure, and honeycomb. Rendering takes approximately 50 ms per frame at 1280 × 720 resolution and one sample per pixel (1 spp), using per-asset decoders with hidden width 64 on an NVIDIA RTX A6000 GPU. All assets are generated from text prompts except the robot, which is reconstructed from synthetically rendered images. Right: close-ups of the feather edge, moth, spider web, and fur, corresponding to the red boxes.}
  \label{fig:teaser}
  \end{minipage}\par
  \vspace{20pt plus 2pt minus 4pt}
}
\makeatother
\maketitle

\begin{abstract}
Generating realistic 3D assets remains constrained by classical representations: surface meshes and compact physically based materials struggle to express objects whose appearance arises from fine geometry or complex light transport, while existing Gaussian-splatting generators typically bake fixed illumination into appearance. We introduce VideoRNG, a two-stage method that generates complete, relightable 3D assets from a text prompt or a single image without prescribing a mesh or analytic material model. A video generator fine-tuned from a pretrained video diffusion backbone first produces a structured RGBA turntable sequence under a known pose, camera, and illumination. A feed-forward large reconstruction model initialized from a pretrained video diffusion backbone then maps the video-VAE latent frames of selected turntable frames directly to millions of anisotropic Gaussians with learned neural appearance features. A shared directional decoder and stochastic Gaussian ray tracer turn this representation, which we call ray-traced Relightable Neural Gaussians (rtRNG), into radiance under novel viewpoints and illumination. To preserve high-frequency appearance while learning geometry, we introduce depth-anchored appearance training, and we make million-Gaussian reconstruction practical through foreground token selection and cost-aware distributed sampling. VideoRNG generates complex assets such as fur, feathers, and spider webs that can be interactively rendered, relit, and integrated into scenes.
\end{abstract}


\section{Introduction}

Creating complex, realistic 3D objects is essential to populating convincing virtual worlds, but shape modeling and material authoring remain labor-intensive and require substantial expertise, which limits the scale and diversity of virtual environments that can be built within a given budget. Generative models may offer a way forward: recent text- and image-conditioned models can rapidly synthesize high-quality 3D assets~\citep{trellis2,xiang2025structured,lin2025diffsplat}. Yet their output remains limited by the representations they produce.

Existing systems generally produce either surface meshes with physically based rendering (PBR) materials or Gaussian radiance fields that bake specific illumination into appearance. PBR materials typically combine diffuse and microfacet specular lobes controlled by textures such as base color, roughness, and metallicity~\citep{Walter2007,Burley2012}. This representation is compact, interpretable, and well supported by existing tools, but a poor fit for objects whose appearance arises from fine geometry or complex light transport, such as fur, feathers, moss, dense foliage, and spider webs. In contrast, image and video models need not explain their output through a prescribed geometric primitive or analytic reflectance model, and can readily depict such visual complexity.

Two lines of work suggest a route beyond this limitation. Gaussian representations can describe geometry too awkward for conventional surface meshes~\citep{3dgs,threedgrt2024}, and can be predicted directly from sparse images~\citep{gslrm2024}, but their stored appearance is commonly entangled with fixed illumination. Neural appearance models instead learn complex, relightable material response without committing to an analytic BRDF~\mbox{\citep{rna2023,8dna}}. However, these methods assume that the underlying mesh geometry is known. Relightable Neural Gaussians (RNG) combine the two ideas by attaching neural directional appearance to Gaussian geometry for relighting~\citep{rng2024}. These methods reconstruct a supplied object rather than generate one from a text or image prompt; VideoNeuMat~\citep{videoneumat} demonstrates that a generative video prior can supply observations for neural-material extraction, but considers planar material samples rather than complete 3D objects.

This raises a natural question: can we extract complete, relightable 3D assets---including geometry and materials outside the practical reach of classical representations---from the rich outputs of a generative video model?

We introduce \textbf{VideoRNG}, a two-stage method for generating relightable neural 3D assets from a text prompt or a single image (Figure~\ref{fig:pipeline}). The first is a video generator fine-tuned to produce an RGBA turntable sequence under a prescribed object-pose, camera, and illumination protocol. The second is a feed-forward large reconstruction model (LRM) initialized from a pretrained video backbone and fine-tuned for reconstruction; it maps the video-VAE latent frames of selected turntable frames directly to millions of anisotropic Gaussians in a single forward pass. Each Gaussian stores geometry and a learned appearance feature rather than a fixed color or PBR parameter set. A shared neural decoder conditions that feature on the incoming light and outgoing view directions to predict radiance. Forward rendering and differentiable training use stochastic Gaussian ray tracing to account for visibility and shadows~\citep{sun2025stochastic,xu2026stoch3dgs}. We call this combination of RNG's neural directional appearance and stochastic Gaussian ray tracing \textbf{rtRNG}: ray-traced Relightable Neural Gaussians. Its appearance feature space and decoder are learned jointly with the reconstructor from rendered images, without requiring a dataset of Gaussian assets as supervision. After training, the shared decoder can be distilled once into a smaller network used by every asset. The assets can be interactively ray-traced from novel cameras, relit, and inserted into a scene with full global illumination (Figure~\ref{fig:teaser}); compared with traditional Gaussian splatting, ray tracing has multiple benefits, including clean scene insertion and arbitrary shadow computation under any lighting.

Moving from VideoNeuMat's planar material samples to complete 3D objects introduces substantial reconstruction challenges. We emit one Gaussian along the camera ray of every foreground pixel, allowing the model to preserve fine and disconnected structures without first reducing them to a mesh. To make this high-resolution representation tractable, we remove background tokens before the transformer and group similarly sized objects during distributed training to avoid GPU idling due to load imbalance in a multi-GPU setting. We further observe that joint geometry and appearance learning produces soft appearance, motivating a fixed spatial reference for learning fine material detail. Our depth-anchored objective places Gaussian centers at ground truth depths during training while supervising the depth predictor for inference. Together, these choices let VideoRNG represent objects such as furry coats, fuzzy dandelions, and delicate spider webs while retaining view- and light-dependent appearance. Our contributions are:
\begin{itemize}[leftmargin=*, itemsep=0pt, topsep=0pt]
  \item We introduce VideoRNG, a two-stage pipeline that extracts complete relightable 3D assets from text- or image-conditioned generative video, without constraining their geometry to a mesh or their appearance to a prescribed PBR material model (Section~\ref{sec:method}).
  \item We initialize a foreground-sparse LRM from a pretrained generative video DiT and fine-tune all of its parameters to directly predict Gaussian geometry and neural directional appearance at $1024^2$ resolution. Our efficient training scheme supports variable-length, million-Gaussian outputs, and learns the neural appearance basis from rendered images without requiring a dataset of Gaussian assets as supervision (Section~\ref{sec:recon}; Supp.~Sec.~\ref{sec:efficient}).
  \item We propose depth-anchored appearance learning: during training, we place Gaussian centers at ground truth depths for rendering and supervise predicted depths with a separate depth loss. Rendering losses optimize the remaining Gaussian attributes and the appearance decoder; inference uses predicted depths (Section~\ref{sec:depth}).
\end{itemize}

Section~\ref{sec:exp} evaluates generation, reconstruction from ground truth views, and relighting; the supplementary material compares depth-training strategies and provides detailed protocols and additional comparisons.

\section{Related Work}

\textbf{Text- and image-conditioned 3D generation.} DreamFusion~\citep{poole2023dreamfusion} optimizes a 3D representation via score distillation from a 2D image prior, while MVDream~\citep{shi2024mvdream} provides a multiview-consistent prior. Feed-forward systems pair generated sparse views with large reconstruction models such as LRM, Instant3D, InstantMesh, LGM, GRM, and GS-LRM~\citep{Hong2023LRMLR,li2023instant3d,xu2024instantmesh,tang2024lgm,xu2024grm,gslrm2024}. More recent generators, including TRELLIS, TRELLIS.2, and DiffSplat, directly produce structured 3D latents or Gaussian assets~\citep{xiang2025structured,trellis2,lin2025diffsplat}. LSRM~\citep{li2026lsrm} reconstructs meshes from posed views, with an inverse-rendering extension that predicts BRDF materials. These methods improve speed and fidelity across several output representations, but none provides learned relightable neural appearance for objects with complex light transport.

\textbf{Video and multiview diffusion as 3D priors.} Multiview diffusion models generate consistent observations for optimization or feed-forward reconstruction~\citep{shi2024mvdream,li2023instant3d}, and SV3D adapts latent video diffusion to orbital novel-view synthesis and 3D generation from one image~\citep{sv3d}. Gen3R couples a pretrained video diffusion model with a VGGT-derived geometry latent space to jointly generate images, point clouds, depth, and cameras~\citep{huang2026gen3r}. Diff4Splat pairs a video diffusion model with a latent dynamic reconstruction model to predict explicit deformable Gaussian fields for 4D scene generation~\citep{pan2026diff4splat}. VideoNeuMat extracts planar neural materials from generated video~\citep{videoneumat}. VideoRNG uses video diffusion both to generate structured observations and to initialize a separate reconstruction backbone, which is fine-tuned with its denoising head replaced by a Gaussian prediction head to produce an explicit, renderable asset; this extends VideoNeuMat's planar materials to complete static objects with relightable neural appearance.

\textbf{Neural appearance representations and generation.} Neural materials replace restricted analytic reflectance models with learned directional appearance, supporting effects such as parallax, self-shadowing, and layered scattering~\citep{neumip,Hierarchical,zeltner24}; RNA and 8DNA extend them to complete relightable assets with global light transport but assume known surface geometry~\citep{rna2023,8dna}. Generative Neural Materials synthesizes planar neural materials from image or text conditioning in a shared feature basis~\citep{gnm}. Concurrent work GenerativeShells~\citep{generativeshells2026}, with partially overlapping authors, generates relightable neural appearance within a user-provided bounding mesh and supports animation through mesh deformation, whereas VideoRNG generates complete relightable Gaussian assets, including geometry and appearance, from text or image prompts. VideoRNG learns its appearance basis jointly with a geometry-predicting LRM from rendered images, without a dataset of pre-existing rtRNG assets.

\textbf{Relightable Gaussian assets and generators.} RNG attaches neural directional appearance to Gaussian geometry, relighting objects with hard surfaces and soft boundaries without an analytic shading model, but fits each object independently~\citep{rng2024}. Large Material Gaussian Model and Luce generate relightable Gaussians by predicting explicit PBR quantities such as albedo, roughness, metallicity, and normals~\citep{ye2025mgm,singh2026luce}. RelitLRM generates relit Gaussian appearance for a target environment map rather than relightable assets~\citep{zhang2025relitlrm}. The concurrent work F-RNG has similar goals to ours but different methods: it reconstructs neural Gaussian assets with frozen relighting priors and shades blended features using rasterization~\citep{fu2026frng}. The above works do not currently ship code or models for comparison.

\textbf{Gaussian rendering.} 3D Gaussian Splatting rasterizes anisotropic Gaussians efficiently~\citep{3dgs}, 3D Gaussian Ray Tracing adds primary and secondary rays~\citep{threedgrt2024}, and sorting-free stochastic ray tracing makes transparent Gaussian clouds practical inside conventional path tracers~\citep{sun2025stochastic}, with a differentiable extension that supports reconstruction and relightable per-Gaussian shading with traced visibility~\citep{xu2026stoch3dgs}. rtRNG builds on these renderers for interactive rendering, shadow rays, and integration with conventional geometry.

\section{Method}
\label{sec:method}

\begin{figure*}[t]
  \centering
  \includegraphics[width=\linewidth,trim={0bp 65bp 0bp 70bp},clip]{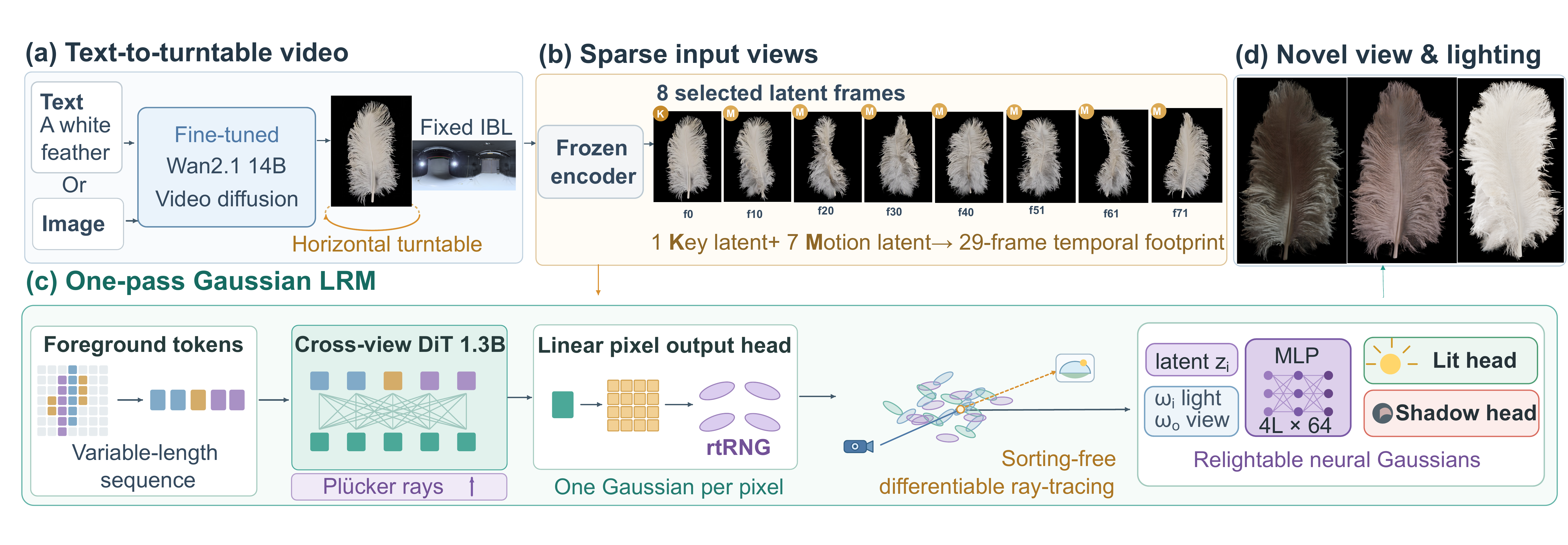}
  \caption{VideoRNG inference pipeline. (a) A fine-tuned video generator produces an 81-frame turntable. (b) 29 uniformly spaced frames are encoded into eight VAE latent frames. (c) The reconstruction DiT processes foreground tokens and predicts one Gaussian per foreground pixel, including a neural appearance feature. The MLP labeled ``4L $\times$ 64'' is the per-asset decoder distilled from the nine-layer, width-512 shared decoder described in Section~\ref{sec:rendering} for final rendering. (d) The resulting asset supports novel views and lights through stochastic Gaussian ray tracing.}
  \label{fig:pipeline}
\end{figure*}

VideoRNG reconstructs relightable assets from text or image (Figure~\ref{fig:pipeline}). The video generator produces a turntable under the prescribed camera trajectory and studio illumination from a text prompt or a single image (a). We encode 29 frames from this standardized sequence into 8 video-VAE latent frames (b), then pass their foreground tokens to a reconstruction model that predicts Gaussian geometry and neural appearance in one forward pass (c). The resulting asset supports novel-view rendering and relighting (d). The inference pipeline is
\begin{equation}
  \begin{aligned}
  \text{Text/Image} &\;\xrightarrow[]{\substack{\text{Finetuned video}\\\text{generator}}}\; \text{Turntable observations} \\
  &\xrightarrow[]{\text{Gaussian LRM}}\; \text{rtRNG asset}.
  \end{aligned}
  \label{eq:pipeline}
\end{equation}
Sections~\ref{sec:obs}--\ref{sec:rendering} describe observation generation, reconstruction, training, and interactive rendering under novel views and lights, respectively. Supp.~Sec.~\ref{sec:efficient} details the sampling procedure and computational cost.

\subsection{Stage 1: Generating Structured Observations}
\label{sec:obs}

\textbf{Generation protocol.} We fine-tune a pretrained latent video diffusion model~\citep{wan2025} with its flow-matching objective on $\sim$22K captioned turntable clips rendered from textured assets. Given a text prompt or a photograph from an arbitrary view that clearly reveals the front, it produces $F{=}81$ turntable frames at $1024\times1024$ resolution. The object completes one rotation about its vertical axis in front of a fixed camera at zero elevation, under studio environment lighting fixed in world coordinates. We selected this lighting for its neutral white balance and clear but not overly sharp specular highlights. Rotation changes both the viewing and illumination directions relative to the object, exposing view-dependent and light-dependent appearance. Joint sequence generation uses the video prior to model cross-frame consistency. The standard variant fine-tunes the pretrained DiT directly; the alpha-capable variant uses the Wan-Alpha VAE and merges its DoRA adapters into the DiT before full fine-tuning~\citep{dong2025wanalpha}. Generated alpha separates foreground from background. Supp.~Sec.~\ref{sec:observation_details} gives the camera calibration and frame selection.

\textbf{Latent observations.} We select $29$ uniformly spaced frames from the turntable, starting at the front view, and encode them with the causal video VAE (temporal compression four, spatial compression eight) into $V{=}8$ latent frames of $16\times128\times128$. One keyframe per latent defines the camera rays used to position the emitted Gaussians, and alpha-derived foreground masks select transformer tokens and emitted Gaussians (details in Supp.~Sec.~\ref{sec:observation_details}).


We also support RGB video on a black background and six independently encoded axis-aligned orthographic views, with the same architecture and depth-anchored objective. RGBA video is most robust overall, but supplementary materials show results from all variants.

\subsection{Stage 2: Reconstructing rtRNG Assets}
\label{sec:recon}

\textbf{Output representation.} An asset comprises $N$ \emph{ray-traced Relightable Neural Gaussians} (rtRNG), $\{g_i\}$, each with a position $\boldsymbol\mu_i$, an anisotropic scale $\mathbf s_i\in\mathbb R^3$, a rotation $\mathbf q_i$, an opacity $\alpha_i$, and a neural appearance latent $\mathbf z_i\in\mathbb R^{32}$. Section~\ref{sec:rendering} describes their rendering.

\textbf{Foreground tokens.} Each latent frame (with resolution $128\times128$) is patchified into $64\times64$ tokens by the patch embedding of a pretrained video diffusion transformer~\citep{wan2025}, which we use as the reconstruction backbone. Let $m_{v,j}\in\{0,1\}$ indicate whether patch $j$ of latent observation $v$ contains any foreground pixel; the transformer processes only
\begin{equation}
  \mathcal T_{\mathrm{fg}} \;=\; \{\, t_{v,j} \;\mid\; m_{v,j}=1 \,\},
  \label{eq:fgtokens}
\end{equation}
and background tokens are removed \emph{before} the first attention layer. Kept tokens retain the rotary embedding of their original latent-frame index and spatial location, so the attention pattern the backbone was pretrained with is preserved on the subset. Each token additionally receives its associated keyframe camera geometry through eight extra input channels of the patch embedding, initialized to zero at the start of training: the six Pl\"ucker coordinates of the pixel ray~\citep{gslrm2024} and the sine and cosine of the keyframe azimuth. No other view or pose information is injected; the full self-attention of the backbone over the kept tokens of all eight latent frames performs the cross-view fusion. On our data the foreground occupies about a fifth of the image area on average ($21.6\%$ of the pixels; $24\%$ of the tokens survive), so this step removes the majority of the tokens and, downstream, the majority of the Gaussians (Supp.~Sec.~\ref{sec:efficient}).

\textbf{Gaussian prediction.} The reconstructor in Figure~\ref{fig:pipeline}(c) fine-tunes all parameters of a pretrained 1.3B-parameter video DiT (30 blocks, width 1536) for 3D reconstruction. We replace the DiT's denoising head with a linear pixel head following GS-LRM~\citep{gslrm2024}. After the final layer normalization and adaptive modulation, a single linear layer maps each retained 1536-dimensional token to a $16\times16$ block of per-pixel attributes. Each token covers a $2\times2$ latent patch, yielding outputs at the full $1024^2$ resolution without a convolutional or VAE decoder. Each attribute vector has 41 used channels: one depth logit, three log-scales, four quaternion components, one opacity logit and the 32-dimensional appearance latent. Every foreground pixel $(v,p)$ emits exactly one Gaussian along its own camera ray. Writing $\mathbf o_{v,p}$ and $\mathbf r_{v,p}$ for the ray origin and unit direction, the head predicts a depth logit $w_{v,p}$ and the position is
\begin{equation}
  \begin{aligned}
  \hat{\boldsymbol\mu}_{v,p} &\;=\; \mathbf o_{v,p} + \hat d_{v,p}\,\mathbf r_{v,p}, \\
  \hat d_{v,p} &\;=\; d_{\min} + (d_{\max}-d_{\min})\,\operatorname{sigmoid}(w_{v,p}),
  \end{aligned}
  \label{eq:raydepth}
\end{equation}
here $[d_{\min},d_{\max}]$ is the depth range of the object's bounding sphere along the ray, and the network predicts the logit $w_{v,p}$ rather than the depth itself. Hats denote quantities obtained from this prediction, while $d^*$ denotes ground truth depth (Section~\ref{sec:depth}). The same parameterization serves perspective and orthographic cameras, differing only in how $\mathbf o$ and $\mathbf r$ are formed. The remaining channels of the attribute vector give the log-scale (exponentiated and clamped), the quaternion (normalized), the opacity (sigmoid), and the 32-dimensional appearance latent. We prune Gaussians whose predicted opacity is below 0.01, which removes low-contribution primitives. The Gaussians from all eight latent observations are concatenated into one asset after opacity pruning, with no additional geometric merging step; at $1024^2$ this is typically one to six million Gaussians per object. Constraining every Gaussian to its associated keyframe camera ray means the network never has to solve a free-space placement problem: geometry is a per-pixel depth, and appearance is a per-pixel latent, both directly aligned with the observation that produced them. 

Rendering cost grows with Gaussian count. We group assets into four buckets by foreground token number and sample each distributed batch from one bucket, reducing cost variation across GPUs while preserving each asset's marginal sampling probability. Supp.~Sec.~\ref{sec:efficient} gives the procedure and training-cost measurements.

\subsection{Training: Depth-Anchored Learning}
\label{sec:depth}

During LRM training, Gaussians from all eight latent observations are rendered jointly under point lights at camera poses distinct from those of the input turntable. Section~\ref{sec:recon} places every Gaussian on its pixel ray at a depth predicted by the network (Eq.~\ref{eq:raydepth}). The reconstructor learns from rendered images of synthetic assets under different cameras and point lights than the input frames, so the natural training strategy is to optimize predicted depth jointly with all other attributes through the rendering loss. When that gradient flows through predicted depth, however, reducing image error can move Gaussian centers as well as change appearance. We observed blurry appearance under such joint training, including when a direct depth loss is added (Supp.~Sec.~\ref{sec:exp_depth}). This motivates a fixed spatial reference for appearance learning: during training, we place Gaussian centers at ground truth depths and supervise the depth predictor separately.

\textbf{Depth anchoring.} Synthetic training assets provide ground truth depths $d^*$ along the input camera rays and reference images $I^*$ under supervision cameras and lights. Let $R(d,\eta)$ denote rendering with depths $d$ and the remaining predicted Gaussian attributes $\eta$ (scales, rotations, opacities, and appearance features). The shared appearance decoder, supervision camera, and lights are implicit in $R$. The network predicts $\hat d$ and $\eta$; the renderer's depth argument $d$ can be supplied by either $\hat d$ or $d^*$. Let $\mathcal L_{\mathrm{img}}$ be the image-reconstruction loss and $\mathcal L_d$ the supervised depth loss. The equations below show these two terms; Supp.~Sec.~\ref{sec:training_details} gives the complete objective, including auxiliary losses. Joint training with depth supervision uses
\begin{equation}
  \mathcal L_{\mathrm{joint}} \;=\; \mathcal L_{\mathrm{img}}\big(R(\hat d,\eta),\,I^*\big)
  \;+\; \lambda_d\,\mathcal L_d(\hat d,\,d^*).
  \label{eq:joint}
\end{equation}
Depth supervision constrains the predicted depths, but the rendering loss can still move Gaussian centers through $\hat d$. We therefore use \emph{depth anchoring} to give appearance learning a fixed spatial reference: the renderer uses ground truth depths $d^*$ to place the centers, while $\hat d$ continues to receive direct depth supervision:
\begin{equation}
  \mathcal L_{\mathrm{ours}} \;=\; \mathcal L_{\mathrm{img}}\big(R(d^*,\eta),\,I^*\big)
  \;+\; \lambda_d\,\mathcal L_d(\hat d,\,d^*).
  \label{eq:ours}
\end{equation}
Equations~\ref{eq:joint} and~\ref{eq:ours} differ in the depths supplied to the training renderer; their depth-supervision term is identical. Under Eq.~\ref{eq:ours}, Gaussian centers are fixed for each training asset, while scales, rotations, opacities, appearance features, and the shared decoder remain trainable. The depth and attribute predictions share a backbone, so anchoring separates the gradient paths through the renderer, not the model parameters. Table~\ref{tab:nodepth_full} evaluates three training variants (rendering loss only, joint depth supervision, and depth anchoring) under novel views and lighting, using predicted depths. Depth anchoring reaches $29.22$\,dB versus $25.39$\,dB with joint depth supervision. Figure~\ref{fig:nodepth_full} shows two representative examples (Supp.~Sec.~\ref{sec:exp_depth}).

\begin{table}[t]
  \centering\footnotesize\setlength{\tabcolsep}{5pt}
  
  \caption{Depth-training ablation. Models train at $256^2$; full-frame PSNR (dB), SSIM and LPIPS (VGG) use predicted depths under novel views and lighting.}
  \label{tab:nodepth_full}
  \begin{tabular}{@{}lrrr@{}}
    \toprule
    Method & PSNR$\uparrow$ & SSIM$\uparrow$ & LPIPS$\downarrow$ \\
    \midrule
    Rendering loss only& 24.49 & 0.759 & 0.162 \\
    Rendering loss $+$ depth loss & 25.39 & 0.779 & 0.148 \\
    Ours, depth-anchored& \textbf{29.22} & \textbf{0.863} & \textbf{0.064} \\
    \bottomrule
  \end{tabular}
\end{table}

\begin{figure}[t]
  \centering\setlength{\tabcolsep}{1.5pt}\small\sffamily
  \newcommand{\nw}{0.232\linewidth}
  \newcommand{\nhead}[1]{\makebox[\nw][c]{\begingroup\footnotesize\setbox0=\hbox{\shortstack{#1}}\parbox[c][20pt][c]{\wd0}{\centering #1}\endgroup}}
  \newcommand{\nc}[1]{\includegraphics[width=\nw]{figs/nodepth/#1.jpg}}
  \begin{tabular}{@{}cccc@{}}
    \nhead{Ground truth} & \nhead{Rendering loss\\only} & \nhead{Rendering loss\\+ depth loss} & \nhead{Ours\\(depth-anchored)} \\
    \nc{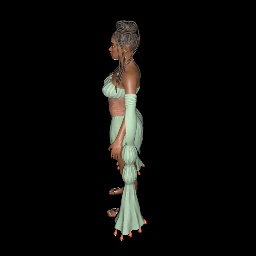} & \nc{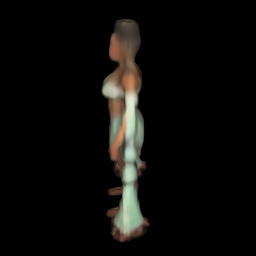} & \nc{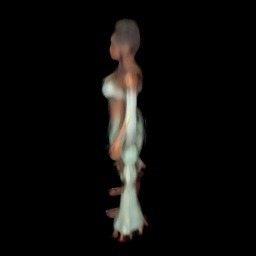} & \nc{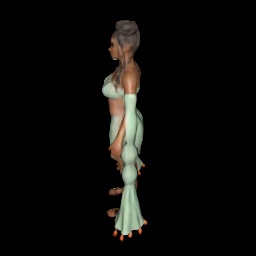} \\
    \nc{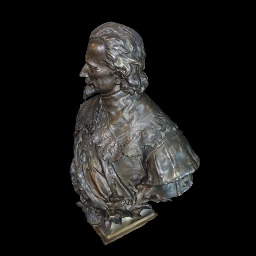} & \nc{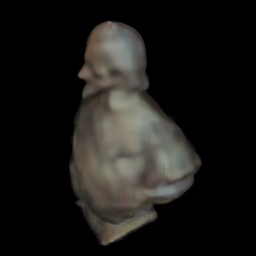} & \nc{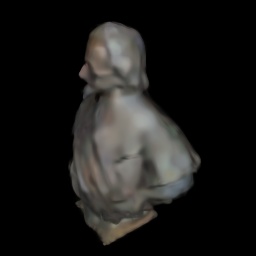} & \nc{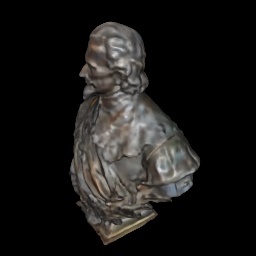} \\
  \end{tabular}
  \caption{Two representative examples from Table~\ref{tab:nodepth_full}, showing assets rendered with predicted depths under novel views and lighting.}
  \label{fig:nodepth_full}
\end{figure}

\textbf{Alpha supervision.} So far the opacities of the Gaussians are only implicitly learned through the $\mathcal L_{\mathrm{img}}$ term, where the final color is $\alpha$-composited based on the individual Gaussians' opacities along the ray. We observed that having a constant black background can confuse the model into reconstructing black silhouettes. To avoid this ambiguity between opacity and color in $\mathcal L_{\mathrm{img}}$, we use a random gray background per sample in the ground truth RGB image, and $\alpha$-composite our Gaussians with the same background. Moreover, we use an additional loss term that supervises the accumulated opacity of the Gaussians along the rays $\hat\alpha$ using the ground truth $\alpha^*\in[0,1]$,
\begin{equation}
  \begin{aligned}
  \mathcal L_{\mathrm{total}} &\;=\; \mathcal L_{\mathrm{img}}\big(R(d^*,\eta),\,I^*\big) \;+\; \lambda_d\,\mathcal L_d(\hat d,\,d^*) \\
  &\quad + \lambda_{\alpha}\,\mathcal L_{\alpha}(\hat\alpha,\,\alpha^*) + \mathcal L_{\mathrm{reg}}.
  \end{aligned}
  \label{eq:total}
\end{equation}
The $\mathcal L_{\mathrm{img}}$ term combines $\ell_2$, VGG LPIPS~\citep{zhang2018lpips}, and $1-\mathrm{SSIM}$~\citep{wang2004ssim}, and together with the soft-alpha silhouette loss ($\mathcal L_{\alpha}$) and foreground depth supervision ($\mathcal L_d$) forms our loss function. Additionally, $\mathcal L_{\mathrm{reg}}$ applies weak binarization regularization on rendered alpha and Gaussian opacity. Supp.~Sec.~\ref{sec:training_details} gives the complete loss and regularizer definitions, weights, and augmentation settings.

\textbf{Inference.} ground truth depth is used only during training on our synthetic data. At inference, for a generated or supplied observation sequence, the reconstructor predicts $\hat d$ and $\eta$, places Gaussian centers with Eq.~\ref{eq:raydepth}, and renders $R(\hat d,\eta)$ under the requested camera and lighting; the change from $R(d^*,\eta)$ to $R(\hat d,\eta)$ is the depth-related train--test mismatch.

\subsection{Rendering rtRNG Assets}
\label{sec:rendering}

A shared decoder $f_\psi(\mathbf z_i,\boldsymbol\omega_{\mathrm{in}},\boldsymbol\omega_{\mathrm{out}})$ predicts directional appearance at render time. It is a nine-layer MLP with hidden width 512, conditioned on the appearance feature and the incoming light and outgoing view directions. (For final rendering, we use per-asset distillation into a four-layer MLP of width 64.) Both directions are defined in the Gaussian's local coordinate frame, before applying its rotation $\mathbf q_i$; world-space directions are mapped into this frame by $\mathbf R(\mathbf q_i)^{\top}$. The directional input comprises the two in-plane components of each direction, Fourier-encoded with six frequencies; the in-plane components of their half vector, encoded with one frequency; and the two components along the Gaussian's third local axis. The rotation that orients the Gaussian also fixes the frame in which its appearance is expressed, so the encoding depends only on the light and view directions relative to the Gaussian, not on any world axis, and anisotropic reflectance can be represented. We adopt the two-output shading formulation ("shadow hint") of RNA~\citep{rna2023}: the decoder predicts two radiance responses for each incident-light and outgoing-view direction pair: $f^{\mathrm{lit}}_\psi$ for a visible incident direction and $f^{\mathrm{shd}}_\psi$ for a self-occluded one. We introduce an improvement: our renderer does not treat this as a binary decision but blends them with the transmittance $T_{i\ell}\in[0,1]$ traced from the Gaussian along the world-space direction $\mathbf R(\mathbf q_i)\boldsymbol\omega_\ell$, where $\boldsymbol\omega_\ell$ is the sampled incident direction in the Gaussian's local frame:
\begin{equation}
  \begin{aligned}
  \mathbf c_i(\boldsymbol\omega_{\mathrm{out}}) &\;=\; \sum_{\ell}
  E_\ell\Big[\,T_{i\ell}\, f^{\mathrm{lit}}_\psi(\mathbf z_i,\boldsymbol\omega_\ell,\boldsymbol\omega_{\mathrm{out}}) \\
  &\quad + (1-T_{i\ell})\, f^{\mathrm{shd}}_\psi(\mathbf z_i,\boldsymbol\omega_\ell,\boldsymbol\omega_{\mathrm{out}})\Big],
  \end{aligned}
  \label{eq:shade}
\end{equation}
where $E_\ell$ is the weighted contribution of incident-light sample $\ell$ at the Gaussian. The summation accommodates point, area, and environment lighting. In a controlled fitting study the blend improves over the binary selection at every matched training checkpoint, by about $0.9$\,dB late in training (Supp.~Sec.~\ref{app:softblend}). The computed radiances $\mathbf c_i$ are then alpha-composited along camera rays using the sorting-free stochastic estimator of \citet{sun2025stochastic}; shadow rays behind $T_{i\ell}$ are naturally sorting-free (particle kernel and asset conventions: Supp.~Sec.~\ref{app:kernel}). During training, we use the differentiable stochastic formulation of \citet{xu2026stoch3dgs} to propagate image-space losses to the Gaussian parameters and neural appearance. Storing a neural latent per Gaussian follows Relightable Neural Gaussians~\citep{rng2024} but our ray-traced formulation is much simpler to integrate into downstream renderers; we also learn the representation and its shared appearance basis end-to-end during LRM training, and predict a new rtRNG asset in a single forward pass, rather than fit it per object from hundreds of consistent observations.

We integrate rtRNG assets into a path tracer for scene rendering with inter-asset shadowing and global illumination. We support per-asset distillation of the shared decoder $f_\psi$ for faster rendering. Per-asset distillation fits a width-64 decoder to the frozen teacher's directional responses on that asset's Gaussian features, without image supervision or updates to the Gaussian attributes. It takes approximately 2 minutes per object on one GPU. The rendering time in Figure~\ref{fig:teaser} uses these per-asset decoders; Supp.~Sec.~\ref{app:decoder_distillation} details their architecture, sampling, and objective, and Supp.~Sec.~\ref{app:rendering_performance} reports gallery rendering performance.

\begin{table*}[t]
  \centering
  \begin{minipage}[t]{0.59\linewidth}
  \vspace{0pt}
  \centering\footnotesize\setlength{\tabcolsep}{4pt}
  
  \caption{Text-conditioned generation on 105 objects (four views each). Best scores are bold; evaluation details are in Supp.~Sec.~\ref{sec:text3d}.}
  \label{tab:text3d_text}

  \begin{tabular}{@{}l ccccc@{}}
    \toprule
    Method & CLIP-T$\uparrow$ & R@1$\uparrow$ & R@5$\uparrow$ & PickScore$\uparrow$ & ImageReward$\uparrow$ \\
    \midrule
    Ours & \textbf{28.1} & \textbf{0.81} & \textbf{1.00} & \textbf{20.59} & \textbf{-0.19} \\
    TRELLIS text-xlarge & 24.1 & 0.56 & 0.91 & 19.70 & -1.00 \\
    LGM (MVDream views) & 25.9 & 0.74 & 0.99 & 19.76 & -0.73 \\
    DiffSplat (PixArt-$\Sigma$) & 24.3 & 0.59 & 0.91 & 19.55 & -0.90 \\
    Shap-E & 17.6 & 0.12 & 0.35 & 18.32 & -2.11 \\
    GaussianAnything & 19.5 & 0.20 & 0.51 & 18.51 & -1.82 \\
    Direct2.5 & 21.4 & 0.43 & 0.60 & 18.86 & -1.53 \\
    \bottomrule
  \end{tabular}
  \end{minipage}\hfill
  \begin{minipage}[t]{0.39\linewidth}
  \vspace{0pt}
  \centering\footnotesize\setlength{\tabcolsep}{4pt}
  
  \caption{Relighting on the same 50 complex validation objects as Table~\ref{tab:gtcmp_main}, under three novel environment maps. Best values are bold.}
  \label{tab:relight_gt}
  \begin{tabular}{@{}l ccc@{}}
    \toprule
    Method & PSNR$\uparrow$ & SSIM$\uparrow$ & LPIPS$\downarrow$ \\
    \midrule
    Ours & \textbf{25.47} & \textbf{0.870} & \textbf{0.097} \\
    LSRM & 22.53 & 0.834 & 0.125 \\
    Pixal3D & 22.30 & 0.828 & 0.105 \\
    Pixal3D (multi-view) & 21.50 & 0.812 & 0.126 \\
    ReLi3D & 21.21 & 0.817 & 0.177 \\
    TRELLIS.2-4B & 20.68 & 0.793 & 0.154 \\
    \bottomrule
  \end{tabular}

  \end{minipage}
\end{table*}

\begin{figure*}[t]
  \begin{minipage}[t]{0.48\linewidth}
  \vspace{0pt}
  \centering
    \centering
    \includegraphics[width=\linewidth]{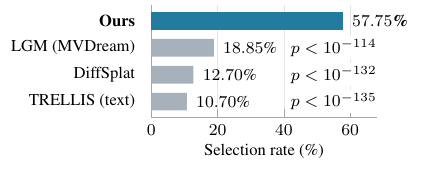}
    \caption{Human preference for text-to-3D renderings, judging visual quality and text alignment jointly. Annotations show $p$-value bounds for paired tests of per-participant selection rates against VideoRNG (Supp.~Sec.~\ref{sec:text3d}).}
    \label{fig:user_study}
\end{minipage}\hfill
\begin{minipage}[t]{0.49\linewidth}
  \vspace{0pt}
    \makeatletter\def\@captype{table}\makeatother
    \centering\footnotesize\setlength{\tabcolsep}{2pt}
    \caption{GT-view reconstruction on 50 complex validation objects. Hunyuan3D-2mv and Pixal3D receive four input views; LSRM and ReLi3D sixteen posed views. The two lighting settings use different evaluation cameras (Supp.~Sec.~\ref{sec:gtcmp}).}
    \label{tab:gtcmp_main}
    \begin{tabular}{@{}lcccc@{}}
      \toprule
      & \multicolumn{2}{c}{Input light} & \multicolumn{2}{c}{New light} \\
      \cmidrule(lr){2-3}\cmidrule(lr){4-5}
      Method & PSNR$\uparrow$ & LPIPS$\downarrow$ & PSNR$\uparrow$ & LPIPS$\downarrow$ \\
      \midrule
      Ours  & \textbf{26.25} & \textbf{0.138} & \textbf{25.99} & \textbf{0.154} \\
      TRELLIS image-large (multi-image)  & 21.20 & 0.264 & 20.54 & 0.284 \\
      LGM  & 21.73 & 0.227 & 21.44 & 0.248 \\
      Hunyuan3D-2mv + Paint  & 19.82 & 0.269 & 19.80 & 0.288 \\
      TRELLIS.2-4B  & 19.42 & 0.268 & 20.04 & 0.280 \\
      LSRM  & 18.95 & 0.192 & 20.52 & 0.192 \\
      Pixal3D  & 21.36 & 0.193 & 21.96 & 0.207 \\
      ReLi3D  & 18.35 & 0.251 & 19.21 & 0.265 \\
      \bottomrule
    \end{tabular}
\end{minipage}
\end{figure*}

\begin{figure*}[t]
  \centering\setlength{\tabcolsep}{0pt}\small\sffamily
  \newcommand{\cw}{0.11875\linewidth}
  \newcommand{\cc}[1]{\cellcolor{black}\includegraphics[width=\cw]{figs/cmp/#1.jpg}}
  \newcommand{\fc}[1]{\cellcolor{black}\begingroup\setbox0=\hbox{\includegraphics{figs/text3d/#1.jpg}}\includegraphics[width=\cw,trim={0.84048\wd0 0.0117\ht0 0.0019\wd0 0.0117\ht0},clip]{figs/text3d/#1.jpg}\endgroup}
  \newcommand{\na}{{\setlength{\fboxsep}{0pt}\colorbox{black}{\parbox[b][\cw][c]{\cw}{\centering\color{white}\tiny n/a}}}}
  \newcommand{\cmpHead}[1]{\makebox[\cw]{#1}}
  \newcommand{\cmpRow}[1]{\hbox to1.2em{\hfil\vbox to\cw{\vfil\hbox{\rotatebox{90}{\small #1}}\vfil}\hfil}}
  \begin{tabular}{@{}c@{\hspace{2pt}}*{8}{c}@{}}
     & \cmpHead{Generated} & \cmpHead{\textbf{Ours}} & \cmpHead{Pixal3D} & \cmpHead{TRELLIS.2} & \cmpHead{TRELLIS v1} & \cmpHead{DiffSplat} & \cmpHead{TRELLIS text} & \cmpHead{LGM} \\
    \vbox{\offinterlineskip\cmpRow{web}\cmpRow{fur coat}\cmpRow{feather}} & \multicolumn{8}{@{}c@{}}{\includegraphics[width=0.95\linewidth]{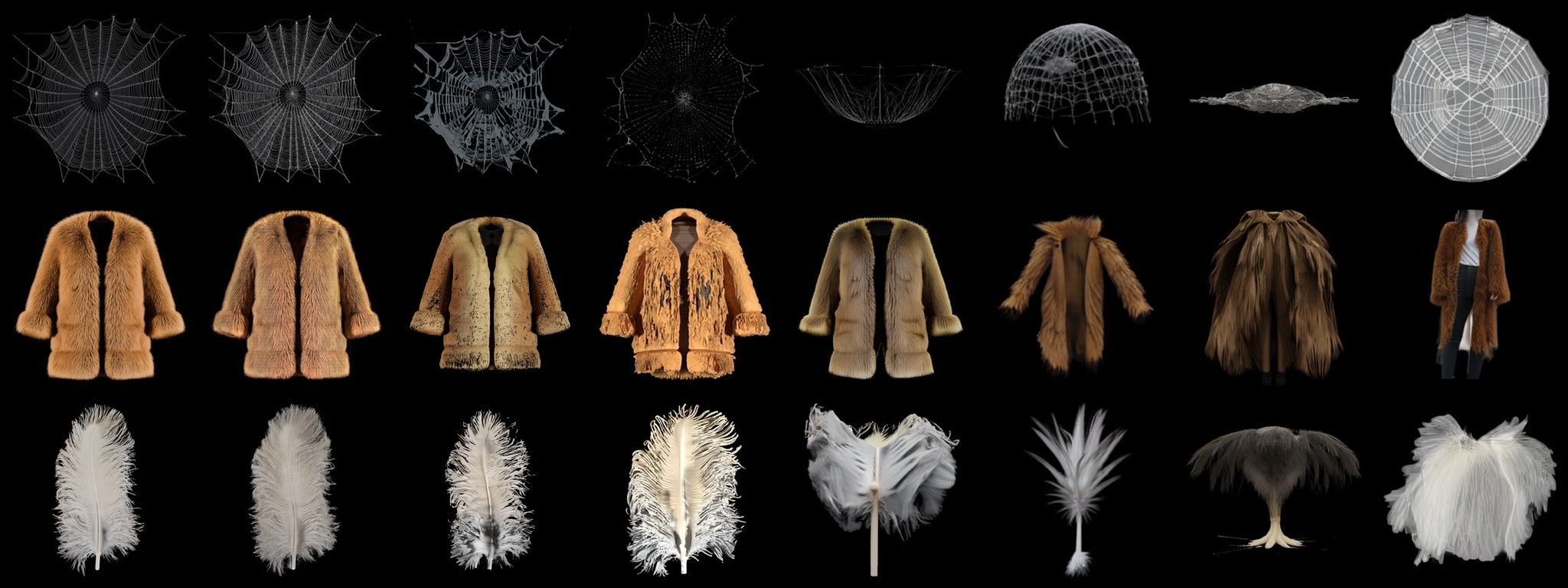}} \\
  \end{tabular}
  \caption{Comparison using generated inputs, not ground truth. Ours and Pixal3D use generated views; TRELLIS.2 and TRELLIS v1 use the shown front frame; DiffSplat, TRELLIS text, and LGM (MVDream) use text prompts.}
  \label{fig:compare}
\end{figure*}

\section{Experiments}
\label{sec:exp}

\subsection{Experimental Setup}
\label{sec:exp_setup}

\textbf{Data and training.} We train the video generator and the large reconstruction model (LRM) on approximately 22K textured assets curated from TexVerse~\citep{texverse}. We first discard assets with overly simple PBR materials, then keep those whose captions describe materials, preferring complex materials and single objects rather than full scenes. For each asset, we render an 81-frame RGBA turntable at $1024^2$ under a fixed environment light, following Section~\ref{sec:obs}. This set is used to train our video generator and provides the input observations to the LRM. To supervise the LRM, we also render 128 views per asset under different camera/point lights than the 81-frame trajectory, with ground truth depth and masks. This set contains 64 frames with fixed and structured camera/light trajectories along with 64 per-asset random camera/light positions, with the aim of providing more robustness to the LRM supervision. Reconstruction benchmarks evaluate validation objects under held-out views and lights; text-generation benchmarks evaluate assets synthesized from prompts.

The two models are trained separately. We fine-tune the video generator on the rendered turntable clips with its flow-matching objective for three epochs. The LRM is initialized from the pretrained 1.3B video DiT and fine-tuned end to end with the depth-anchored objective (Section~\ref{sec:depth}), following a coarse-to-fine schedule from $256^2$ to $1024^2$ with AdamW at a constant learning rate of $2\times10^{-5}$. Supp.~Sec.~\ref{sec:training_details} reports the step counts, compute setup, and input augmentation for both models.

\textbf{Baselines.} We use each method's supported conditioning. Due to different designs, we cannot simply match the number of views provided to the methods. In Figure~\ref{fig:compare}, VideoRNG receives generated multi-view observations; Pixal3D~\citep{li2026pixal3d} receives four; TRELLIS.2~\citep{trellis2} and TRELLIS~\citep{xiang2025structured} receive the generated front view; and DiffSplat~\citep{lin2025diffsplat}, TRELLIS text-xlarge, and LGM with MVDream~\citep{tang2024lgm,shi2024mvdream} receive text prompts. All outputs are rendered with the same orthographic cameras. Supp.~Secs.~\ref{sec:text3d} and~\ref{sec:gtcmp} give the broader text-generation and ground truth reconstruction protocols.

We relight TRELLIS.2 meshes with PBR materials in Blender Cycles; TRELLIS image-large and text-xlarge, LGM, and DiffSplat use Gaussians with baked appearance. In the ground truth benchmark, Hunyuan3D-2mv + Paint produces a textured mesh, relit with its texture as albedo, and LSRM, Pixal3D~\citep{li2026pixal3d} and ReLi3D~\citep{dihlmann2026reli3d} predict meshes with PBR materials, which we use for relighting. VideoRNG predicts Gaussians with neural appearance features: under a new environment map, its shared decoder evaluates responses to sampled incident-light directions, combined with ray-traced visibility as in Eq.~\ref{eq:shade}.


\subsection{Generation and Reconstruction Quality}
\label{sec:exp_main}

We provide extensive qualitative results in the supplementary material: 179 generated asset results across all three input variants, ground-truth reconstruction and relighting comparisons on 15 synthetic assets, and five multi-asset scenes.


\textbf{Reconstruction of generated observations.} Figure~\ref{fig:compare} compares generated objects under the same camera, grouping methods by their inputs as described in Section~\ref{sec:exp_setup}. Our reconstruction follows the generated observations; the single-view generators change the web into a bowl and the coat into pale yellow. As expected, the text-only generators produce different objects, but with less visual fidelity: a dome, a flat disc and a round mesh for the web.

\textbf{Generation from prompts.} Figure~\ref{fig:teaser} shows generated assets in one scene; Table~\ref{tab:text3d_text} evaluates text-conditioned generation. CLIP-T~\citep{radford2021clip} measures text--image similarity, and R-Precision (R@$k$) the fraction of objects whose prompt ranks in the CLIP top-$k$ of 105 prompts. PickScore~\citep{kirstain2023pickapic} and ImageReward~\citep{xu2023imagereward} are learned human-preference metrics (Supp.~Sec.~\ref{sec:text3d}). VideoRNG leads all five metrics; LGM is the strongest baseline on each. Text-only methods may generate different instances of the prompted object. 

\textbf{Human preference study.} In a user study with 505 participants, each shown a target description and one rendered image per method, participants choose the result they prefer for visual quality and description alignment jointly. Figure~\ref{fig:user_study} shows that VideoRNG receives a majority of selections (57.75\%), followed by LGM (MVDream) at 18.85\%. Supp.~Sec.~\ref{sec:text3d} provides the survey protocol.

Figure~\ref{fig:text3d_text} in Supp.~Sec.~\ref{sec:text3d} provides a qualitative comparison of text-conditioned generation with the baselines.
Figure~\ref{fig:text3d_image} shows an image-conditioned qualitative comparison with TRELLIS.2-4B, TRELLIS image-large, and LGM with ImageDream, which receive the front view generated by our text-conditioned pipeline, whereas our reconstruction uses multiple generated observations. Supp.~Sec.~\ref{sec:prompt_construction} contains the original prompt-construction instructions.


\textbf{Reconstruction accuracy comparison.} To isolate reconstruction quality from errors in generated observations, we supply each method with rendered ground truth views of the object. We evaluate 50 complex validation objects on eight random perspective views, each under a different random HDRI. Table~\ref{tab:gtcmp_main} reports PSNR and LPIPS; Supp.~Sec.~\ref{sec:gtcmp} provides the full protocol and SSIM results (Table~\ref{tab:gtcmp}). VideoRNG has the best reported PSNR, SSIM, and LPIPS in both settings. Under new illumination, each method is rendered as described in Section~\ref{sec:exp_setup} and Supp.~Sec.~\ref{sec:gtcmp}; Figure~\ref{fig:gtcmp_a_hv} compares reconstruction under novel views and lighting, including relightable baselines LSRM-PBR~\citep{li2026lsrm}, ReLi3D~\citep{dihlmann2026reli3d}, and Pixal3D~\citep{li2026pixal3d}; quantitative results are reported in Tables~\ref{tab:gtcmp_main} and~\ref{tab:gtcmp}. Figure~\ref{fig:gtcmp_single} and Table~\ref{tab:gtcmp_single} compare full single-image pipelines.

\textbf{Real-image inputs.} Figure~\ref{fig:real_compare} compares our method with seven single-image baselines on six real photographs of objects with fine structure, including lace, openwork, and carved relief.

\textbf{View and illumination generalization.} We test whether reconstructed appearance remains consistent under novel views and rotation of the known studio environment map, comparing 20 synthetic validation assets with Blender ground truth (Figure~\ref{fig:relight}). Mean PSNR is $29.26$\,dB on the input frames, $29.11$\,dB on the novel orbit, and $29.09$\,dB under new illumination. The moving highlights and darkening of surfaces follow the ground truth. Table~\ref{tab:gtcmp_main} evaluates new HDRIs. We further compare relighting against relightable baselines on the same 50 complex validation objects under three novel environment maps (Table~\ref{tab:relight_gt}). Figure~\ref{fig:relight_cmp} provides separate qualitative comparisons on generated objects and gallery objects with ground truth.

\textbf{Direct per-object optimization.} As expected, per-object fitting improves input-view PSNR while breaking novel view / lighting performance (Figure~\ref{fig:directopt}). Our feed-forward LRM better preserves fine structure under novel views and illumination. Supp.~Sec.~\ref{sec:directopt} gives the setup and comparison.

\section{Conclusion}

\textbf{VideoRNG} turns text or image prompts into relightable \textbf{rtRNG} assets through video generation and feed-forward reconstruction. Shared neural appearance captures fine structures for relighting and scene integration. Depth anchoring improves PSNR over joint depth supervision ($29.22$ versus $25.39$\,dB).




\section*{LLM Usage Statement}

Claude and OpenAI Codex were used in code development and experiment execution, to construct candidate object descriptions for text-conditioned generation, and to assist with manuscript editing. Supp.~Sec.~\ref{sec:prompt_construction} includes the original prompt-construction instructions. The authors take full responsibility for the paper’s claims and results.

{
\small
\bibliographystyle{ieeenat_fullname}
\bibliography{iclr2027_conference}
}
\clearpage
\setlength{\textfloatsep}{20pt plus 2pt minus 4pt}
\setlength{\dbltextfloatsep}{20pt plus 2pt minus 4pt}
\raggedbottom
\maketitlesupplementary
\appendix
\setcounter{figure}{0}
\setcounter{table}{0}
\setcounter{equation}{0}
\renewcommand{\thefigure}{S\arabic{figure}}
\renewcommand{\thetable}{S\arabic{table}}
\renewcommand{\theequation}{S\arabic{equation}}
\renewcommand{\theHfigure}{S\arabic{figure}}
\renewcommand{\theHtable}{S\arabic{table}}
\renewcommand{\theHequation}{S\arabic{equation}}

\renewcommand{\floatpagefraction}{0.75}
\renewcommand{\dblfloatpagefraction}{0.75}
\setcounter{dbltopnumber}{3}
\makeatletter
\setlength{\@fptop}{0pt}
\setlength{\@fpsep}{\floatsep}
\setlength{\@fpbot}{0pt plus 1fil}
\setlength{\@dblfptop}{0pt}
\setlength{\@dblfpsep}{\dblfloatsep}
\setlength{\@dblfpbot}{0pt plus 1fil}
\makeatother
\section{Comparison with text-conditioned 3D generators}
\label{sec:text3d}

We compare our full text-to-3D pipeline with public text-conditioned generators on 105 selected objects. Following the prompt-construction protocol in Supp.~Sec.~\ref{sec:prompt_construction}, this set emphasizes intricate geometry and appearance visible from lateral viewpoints. The baselines are TRELLIS text-xlarge, LGM with MVDream, DiffSplat (PixArt-$\Sigma$), Shap-E~\citep{jun2023shape}, GaussianAnything~\citep{lan2025gaussiananything}, and Direct2.5~\citep{lu2024direct25}, each run with its official code and default settings. All outputs are rendered with the same four-view rig and studio lighting, using Blender for meshes and a 3DGS rasterizer for Gaussians. Text-conditioned methods receive only the prompt and may generate different instances of the described object, so Table~\ref{tab:text3d_text} reports reference-free scores. CLIP-T is ViT-L/14 text--image similarity multiplied by 100~\citep{radford2021clip}. R@$k$ is the fraction of objects whose own prompt ranks among the top-$k$ of the 105 prompt entries by CLIP similarity. Scores are averaged over four rendered side views per object. PickScore~\citep{kirstain2023pickapic} and ImageReward~\citep{xu2023imagereward} are learned human-preference models.

\begin{figure*}[tp]
  \centering\setlength{\tabcolsep}{2pt}\renewcommand{\arraystretch}{0.6}
  \begin{tabular}{@{}p{0.12\linewidth}@{\hspace{3pt}}p{0.85\linewidth}@{}}
    & \parbox[b]{0.1365\textwidth}{\centering\tiny A six-strand braided challah loaf}\hspace{0.0054\textwidth}\parbox[b]{0.1365\textwidth}{\centering\tiny An open Victorian lace parasol with a canopy of\ldots}\hspace{0.0054\textwidth}\parbox[b]{0.1365\textwidth}{\centering\tiny A pyrite crystal cluster on matrix}\hspace{0.0054\textwidth}\parbox[b]{0.1365\textwidth}{\centering\tiny A prickly pear cactus cluster with ripe fruits}\hspace{0.0054\textwidth}\parbox[b]{0.1365\textwidth}{\centering\tiny A giant sequoia exposed root system}\hspace{0.0054\textwidth}\parbox[b]{0.1365\textwidth}{\centering\tiny A large ostrich plume feather} \\[1pt]
    \parbox[b][\dimexpr0.1372\textwidth\relax][c]{0.12\textwidth}{\raggedright\scriptsize Ours (relightable)} & \includegraphics[width=\linewidth]{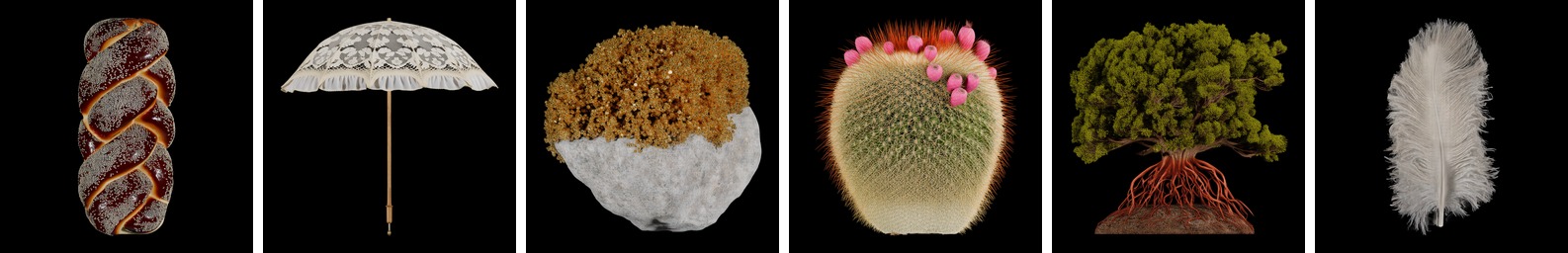} \\
    \parbox[b][\dimexpr0.1372\textwidth\relax][c]{0.12\textwidth}{\raggedright\scriptsize TRELLIS.1 (Gaussian)} & \includegraphics[width=\linewidth]{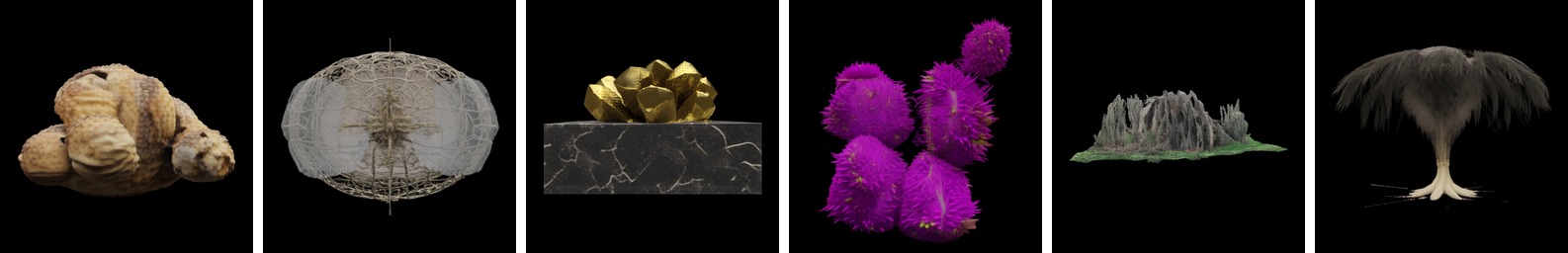} \\
    \parbox[b][\dimexpr0.1372\textwidth\relax][c]{0.12\textwidth}{\raggedright\scriptsize LGM (MVDream)} & \includegraphics[width=\linewidth]{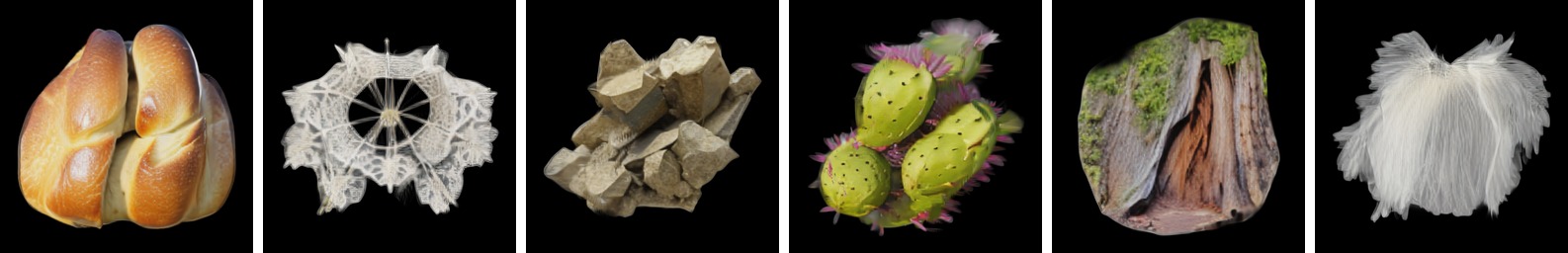} \\
    \parbox[b][\dimexpr0.1372\textwidth\relax][c]{0.12\textwidth}{\raggedright\scriptsize DiffSplat} & \includegraphics[width=\linewidth]{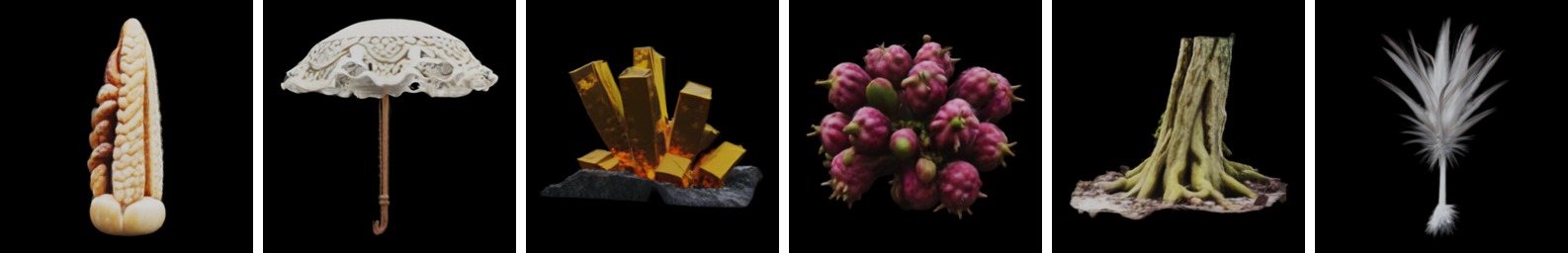} \\
    \parbox[b][\dimexpr0.1372\textwidth\relax][c]{0.12\textwidth}{\raggedright\scriptsize Gaussian\-Anything} & \includegraphics[width=\linewidth]{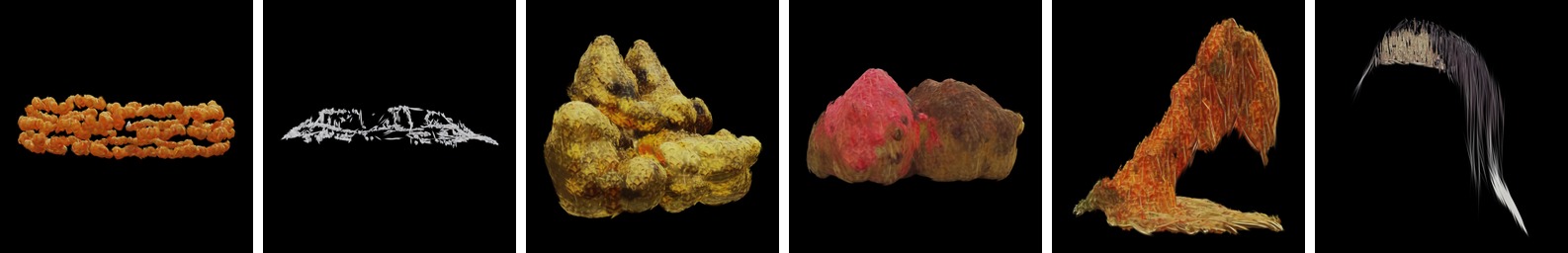} \\
    \parbox[b][\dimexpr0.1372\textwidth\relax][c]{0.12\textwidth}{\raggedright\scriptsize Direct2.5} & \includegraphics[width=\linewidth]{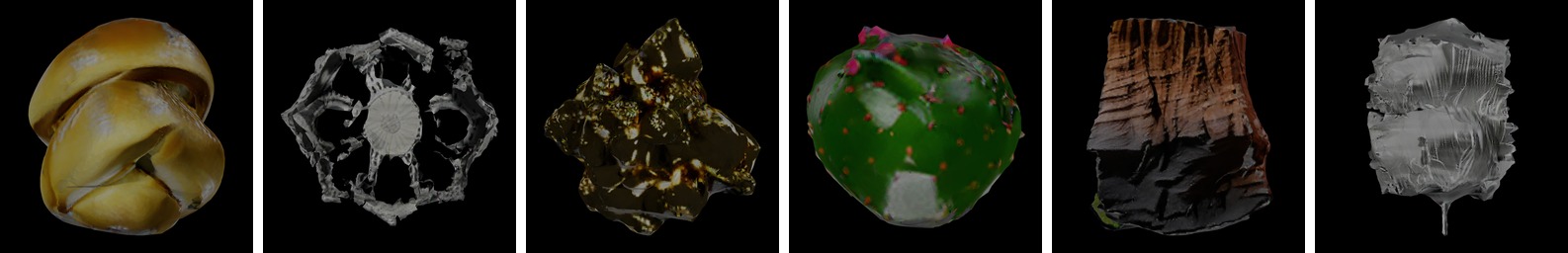} \\
  \end{tabular}
  \caption{Comparison with text-conditioned generation methods on the six objects from Fig.~\ref{fig:text3d_image} (columns; headers summarize the prompt subjects), front views. Our row is the full text pipeline (prompt $\to$ generated views $\to$ LRM). Note that only our method produces relightable assets.}
  \label{fig:text3d_text}
\end{figure*}

\begin{figure*}[tp]
  \centering\setlength{\tabcolsep}{2pt}\renewcommand{\arraystretch}{0.6}
  \begin{tabular}{@{}p{0.12\linewidth}@{\hspace{3pt}}p{0.85\linewidth}@{}}
    \parbox[b][\dimexpr0.1372\textwidth\relax][c]{0.12\textwidth}{\raggedright\scriptsize Input (generated)} & \includegraphics[width=\linewidth]{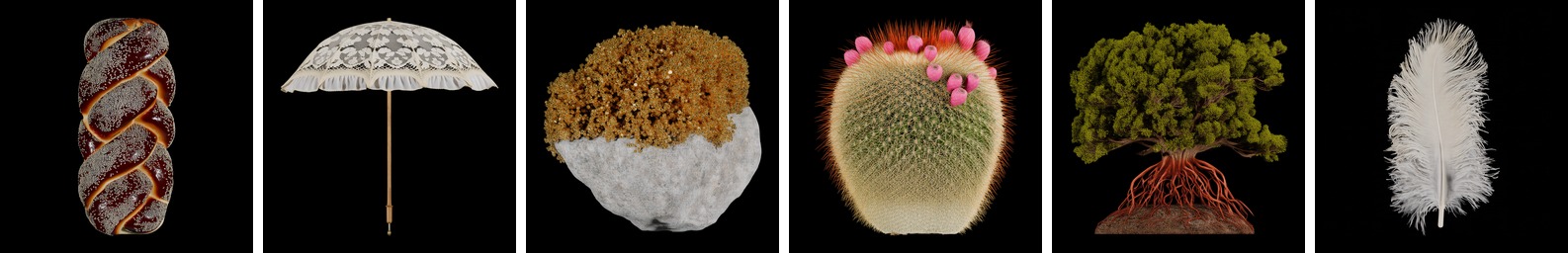} \\
    \parbox[b][\dimexpr0.1372\textwidth\relax][c]{0.12\textwidth}{\raggedright\scriptsize Ours} & \includegraphics[width=\linewidth]{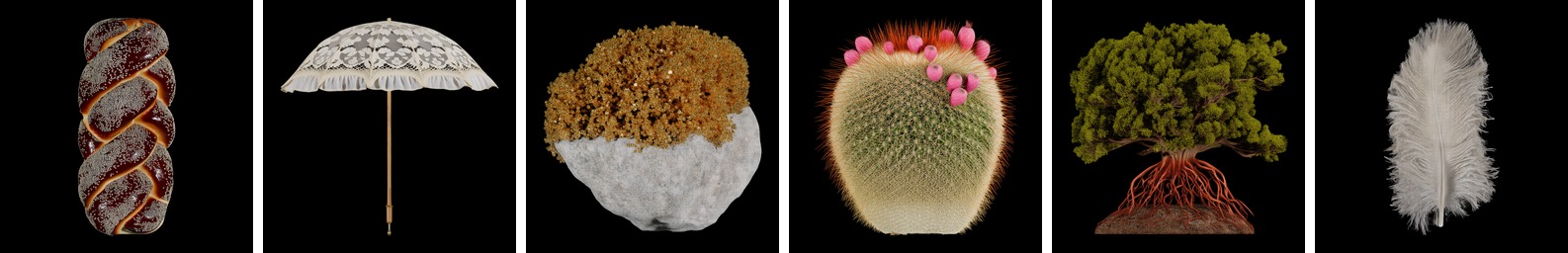} \\
    \parbox[b][\dimexpr0.1372\textwidth\relax][c]{0.12\textwidth}{\raggedright\scriptsize TRELLIS.2-4B} & \includegraphics[width=\linewidth]{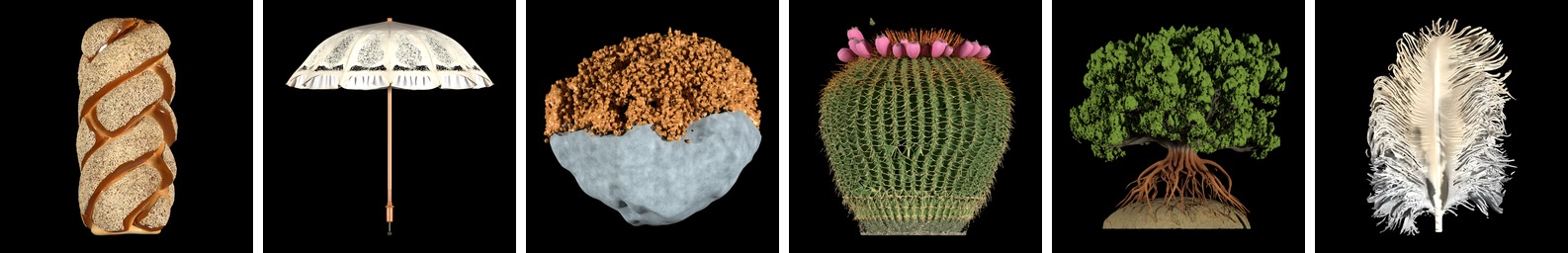} \\
    \parbox[b][\dimexpr0.1372\textwidth\relax][c]{0.12\textwidth}{\raggedright\scriptsize TRELLIS.1 (Gaussian)} & \includegraphics[width=\linewidth]{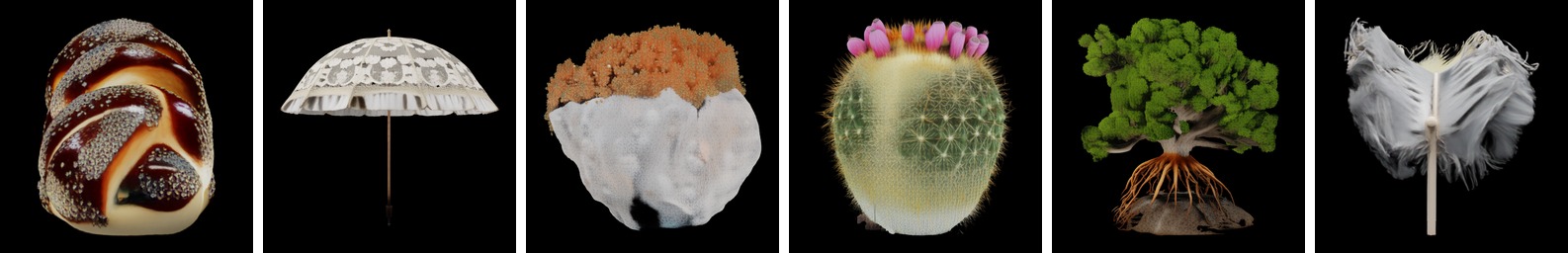} \\
    \parbox[b][\dimexpr0.1372\textwidth\relax][c]{0.12\textwidth}{\raggedright\scriptsize LGM (Image\-Dream)} & \includegraphics[width=\linewidth]{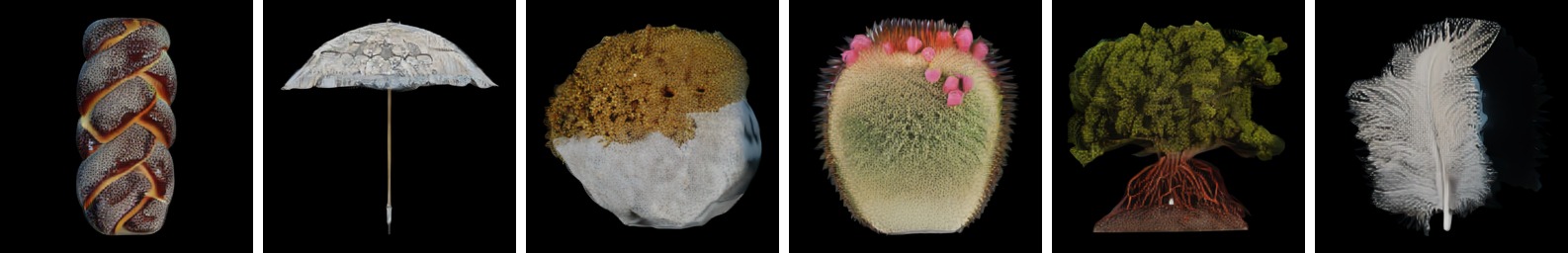} \\
    \parbox[b][\dimexpr0.1372\textwidth\relax][c]{0.12\textwidth}{\raggedright\scriptsize GPT-6 Astra Ultra (Blender asset)} & \includegraphics[width=\linewidth]{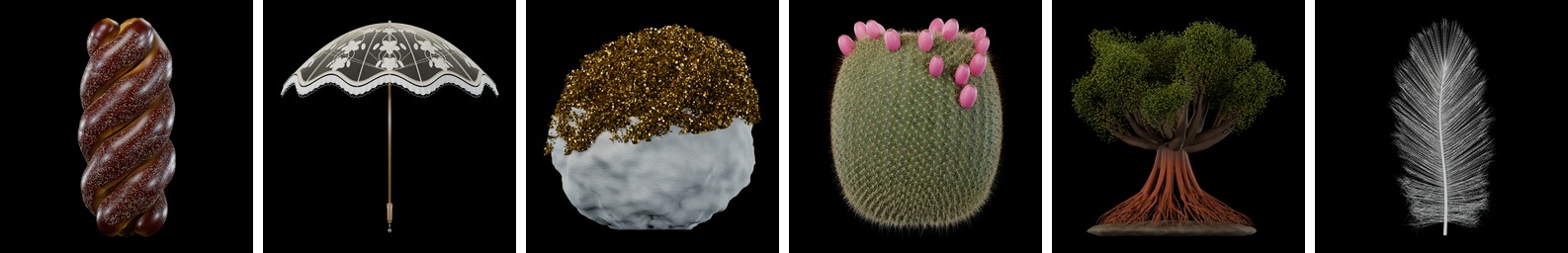} \\
  \end{tabular}
  \caption{Our generated front view used as input to image-conditioned generators. Top row: the view our text pipeline generated for the prompt (same objects and column order as Fig.~\ref{fig:text3d_text}), which TRELLIS.2-4B, TRELLIS image-large and LGM (ImageDream) receive as their input; second row: our reconstruction of the generated views. Front views; each baseline cell is the one of its four side views that best matches the input orientation. Last row: GPT-6 Astra Ultra Blender results, generated from the complete 81-frame video.}
  \label{fig:text3d_image}
\end{figure*}

Figure~\ref{fig:compare} compares the web and fur-coat examples under matched cameras. Figure~\ref{fig:text3d_text} shows six further example objects. Figure~\ref{fig:text3d_image} compares reconstruction from generated observations with image-conditioned generators given our generated front view (TRELLIS.2-4B, TRELLIS image-large, and LGM with ImageDream). For the qualitative GPT-6 Astra comparison~\citep{openai2026astra}, we used Ultra mode and supplied the complete 81-frame generated video with the prompt ``Create a Blender 3D asset that matches this 81-frame video.'' Supp.~Sec.~\ref{sec:gtcmp} separately evaluates reconstruction against ground truth on 50 complex validation objects.



\paragraph{Human preference study.} We evaluate renderings of text-generated assets using 30 single-choice questions. Each question presents a target description prepared for the survey and one rendered image from each of VideoRNG, LGM (MVDream), DiffSplat, and TRELLIS (text). The results are labeled A--D, with method order varying between questions. Respondents make one choice based jointly on overall visual quality and agreement with the target description.

The 505 participants each answered all 30 questions. Figure~\ref{fig:user_study} reports the fraction of choices assigned to each method across all questions and participants. For statistical analysis, we compute one selection rate per method for each participant and compare VideoRNG with each baseline using a two-sided paired $t$-test ($\mathrm{df}=504$). We apply Holm correction to control the probability of making any false-positive significance claim across the three comparisons; all adjusted $p$-values are below $0.001$.

Each question asked: \textquotedblleft{}Considering overall visual quality and how well each result matches the target description, which result do you prefer?\textquotedblright{} The accompanying instructions were:
\begin{quote}
\small
For each object, read the target description and compare results A, B, C, and D. Make one choice considering both overall visual quality (shape, detail, and material appearance) and how well the result matches the described object and its attributes. These descriptions were prepared for this survey. Letters identify positions within each question; the order changes between questions. There are 30 single-choice questions; no name or contact details are requested.
\end{quote}

\begin{figure*}[tp]
  \centering\setlength{\tabcolsep}{0pt}\scriptsize
  \newlength{\rch}\setlength{\rch}{0.1065\textheight}
  \newcommand{\rcRow}[1]{\hbox to1.3em{\hfil\vbox to\rch{\vfil\hbox{\rotatebox{90}{#1}}\vfil}\hfil}}
  \begin{tabular}{@{}c@{}c@{}}
    \vbox{\offinterlineskip\rcRow{Input}\rcRow{\textbf{Ours}}\rcRow{TRELLIS.2}\rcRow{TRELLIS v1}\rcRow{Hunyuan3D-2}\rcRow{Pixal3D}\rcRow{ReLi3D}\rcRow{LGM}\rcRow{DiffSplat}} &
    \includegraphics[height=9\rch]{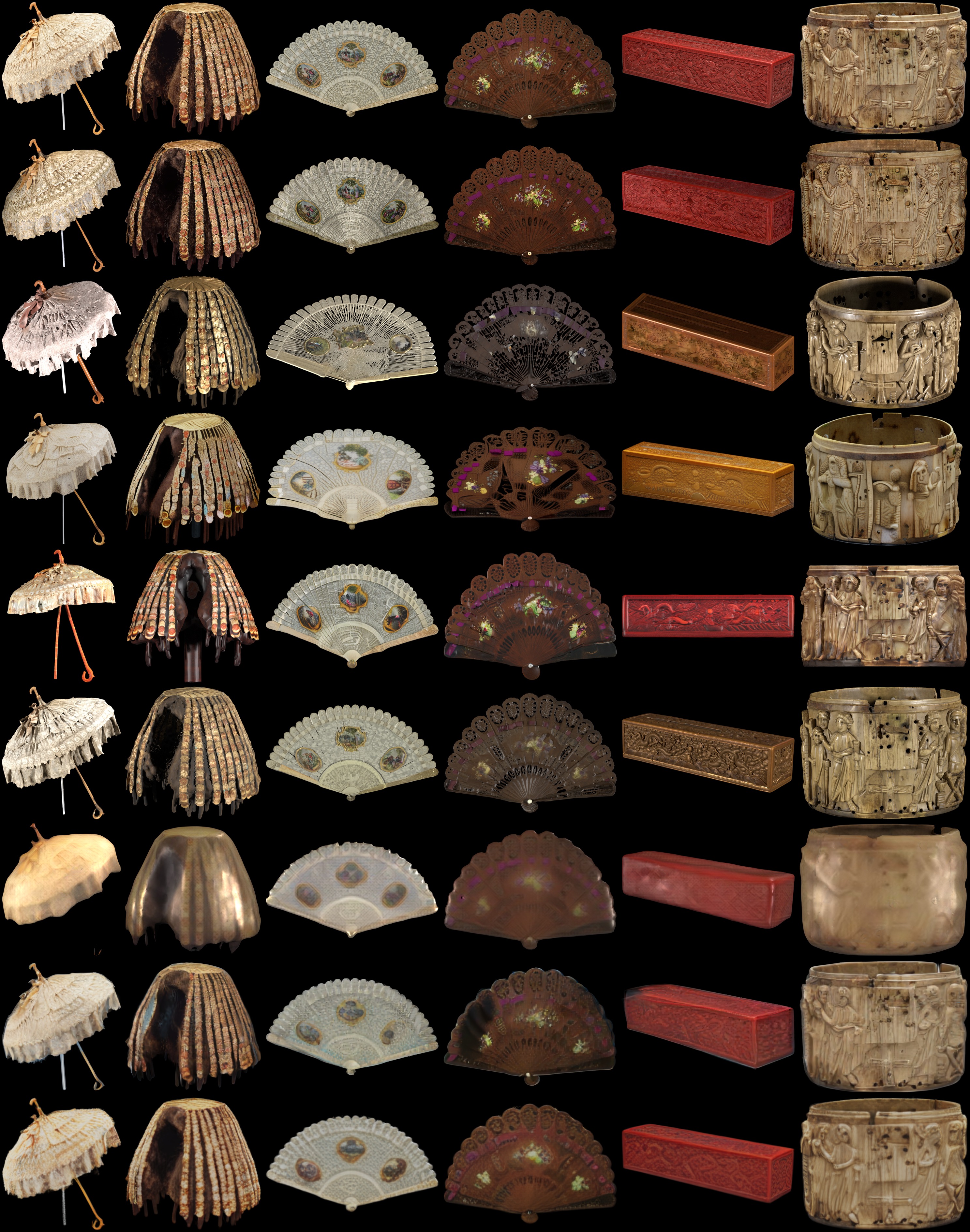} \\
  \end{tabular}
  \caption{Comparison of ours and other methods on real images.}
  \label{fig:real_compare}
\end{figure*}

\section{Comparison against ground truth on complex synthetic objects}
\label{sec:gtcmp}

This section compares our reconstructions with public 3D generators on objects for which exact ground truth exists. We take the 50 most
complex objects of our validation set (TexVerse assets ranked by depth-discontinuity density, silhouette complexity and texture detail of their
rendered views, with foreground coverage between 10\% and 55\%). Two ground truth
sets are used: (i) held-out views under the studio illumination of the training data, and (ii) eight random perspective views of each object, each rendered under a different random HDRI with
random rotation, intensity and color gain. Set (i) tests novel views under the illumination a method has seen in its input, set (ii) novel
views under novel illumination. The two sets use different evaluation views as well as illumination, so their score difference is not a paired comparison of lighting alone.
Baseline outputs are brought into the dataset frame before rendering (orientation and scale chosen by silhouette agreement with the input
views). Mesh baselines are rendered in Blender. Under the novel HDRIs of set~(ii), TRELLIS.2, Pixal3D, LSRM and ReLi3D are relit in Cycles using their PBR materials. TRELLIS image-large and LGM are rendered with a 3DGS rasterizer and retain their baked appearance under new illumination. Our Gaussians are rendered with our own renderer, whose decoder relights them under the per-view HDRI. Public baselines use their official code and default settings, including their own background removal; Hunyuan3D-2mv receives four input views and uses the Hunyuan3D 2.0 shape model followed by Hunyuan3D-Paint~\citep{zhao2025hunyuan3d2}; LSRM~\citep{li2026lsrm} receives sixteen posed perspective views (two rings at $20^\circ$ and $45^\circ$ elevation, rendered under the same illumination) and returns a mesh with PBR materials, run with its released inverse-rendering checkpoint and default settings; Pixal3D~\citep{li2026pixal3d}, the multi-view extension of TRELLIS.2, receives four posed views (front, right, back and left at zero elevation); ReLi3D~\citep{dihlmann2026reli3d} receives sixteen posed perspective views; and the multi-image baseline is TRELLIS image-large~\citep{xiang2025structured} with its official multi-image conditioning on the six input views.

\paragraph{Reconstruction from rendered views.} Every method receives rendered views of the synthetic object. This isolates the reconstruction networks from any generative front end (Table~\ref{tab:gtcmp},
Figure~\ref{fig:gtcmp_a_hv}).

\begin{table*}[tp]
  \centering\footnotesize\setlength{\tabcolsep}{3pt}\renewcommand{\arraystretch}{0.95}
  \setlength{\abovecaptionskip}{4pt}\setlength{\belowcaptionskip}{3pt}
  \caption{Reconstruction from ground truth views on 50 complex validation objects (protocol in Supp.~Sec.~\ref{sec:gtcmp}). Best values are bold.}
  \label{tab:gtcmp}
  \begin{tabular}{@{}l ccc ccc@{}}
    \toprule
     & \multicolumn{3}{c}{Input light} & \multicolumn{3}{c}{New light} \\
    \cmidrule(lr){2-4}\cmidrule(lr){5-7}
    Method & PSNR$\uparrow$ & SSIM$\uparrow$ & LPIPS$\downarrow$ & PSNR$\uparrow$ & SSIM$\uparrow$ & LPIPS$\downarrow$ \\
    \midrule
    Ours & \textbf{26.25} & \textbf{0.829} & \textbf{0.138} & \textbf{25.99} & \textbf{0.797} & \textbf{0.154} \\
    TRELLIS image-large (multi-image) & 21.20 & 0.672 & 0.264 & 20.54 & 0.637 & 0.284 \\
    LGM & 21.73 & 0.698 & 0.227 & 21.44 & 0.675 & 0.248 \\
    Hunyuan3D-2mv + Paint & 19.82 & 0.665 & 0.269 & 19.80 & 0.634 & 0.288 \\
    TRELLIS.2-4B & 19.42 & 0.650 & 0.268 & 20.04 & 0.626 & 0.280 \\
    LSRM & 18.95 & 0.728 & 0.192 & 20.52 & 0.722 & 0.192 \\
    Pixal3D & 21.36 & 0.712 & 0.193 & 21.96 & 0.688 & 0.207 \\
    ReLi3D & 18.35 & 0.698 & 0.251 & 19.21 & 0.683 & 0.265 \\
    \bottomrule
  \end{tabular}
\end{table*}

\begin{figure*}[p]
  \centering\setlength{\tabcolsep}{2pt}\renewcommand{\arraystretch}{0.6}
  \resizebox*{!}{0.9\textheight}{%
  \begin{tabular}{@{}p{0.13\linewidth}@{\hspace{3pt}}p{0.84\linewidth}@{}}
    & \parbox[b]{0.1621\textwidth}{\centering\scriptsize flowering shrub}\hspace{0.0064\textwidth}\parbox[b]{0.1621\textwidth}{\centering\scriptsize mossy stump}\hspace{0.0064\textwidth}\parbox[b]{0.1621\textwidth}{\centering\scriptsize pocket watch}\hspace{0.0064\textwidth}\parbox[b]{0.1621\textwidth}{\centering\scriptsize sewing machine}\hspace{0.0064\textwidth}\parbox[b]{0.1621\textwidth}{\centering\scriptsize wire-wheel tricycle} \\[1pt]
    \parbox[b][\dimexpr0.1629\textwidth\relax][c]{0.13\textwidth}{\raggedright\scriptsize Ground truth} & \includegraphics[width=\linewidth]{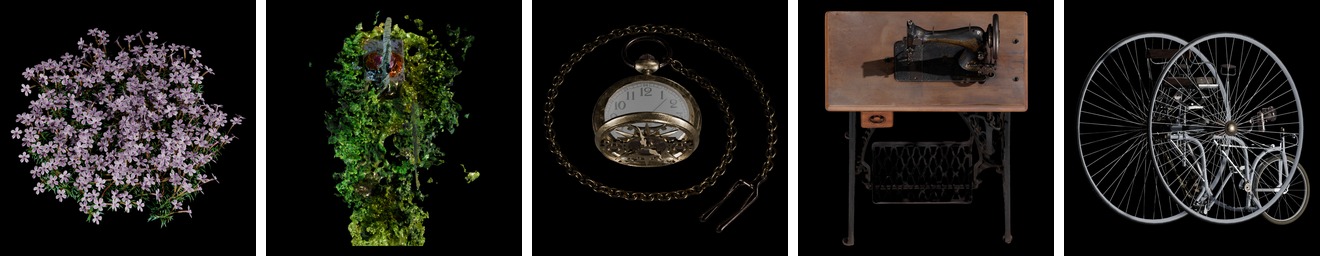} \\
    \parbox[b][\dimexpr0.1629\textwidth\relax][c]{0.13\textwidth}{\raggedright\scriptsize Ours} & \includegraphics[width=\linewidth]{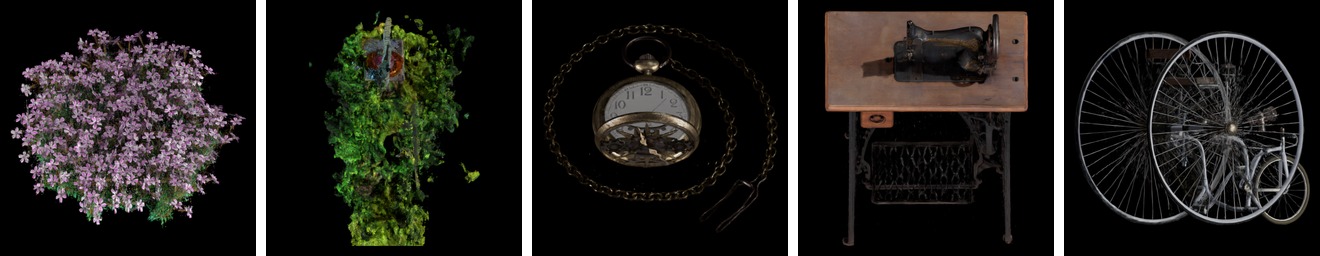} \\
    \parbox[b][\dimexpr0.1629\textwidth\relax][c]{0.13\textwidth}{\raggedright\scriptsize TRELLIS image-large (multi-image)} & \includegraphics[width=\linewidth]{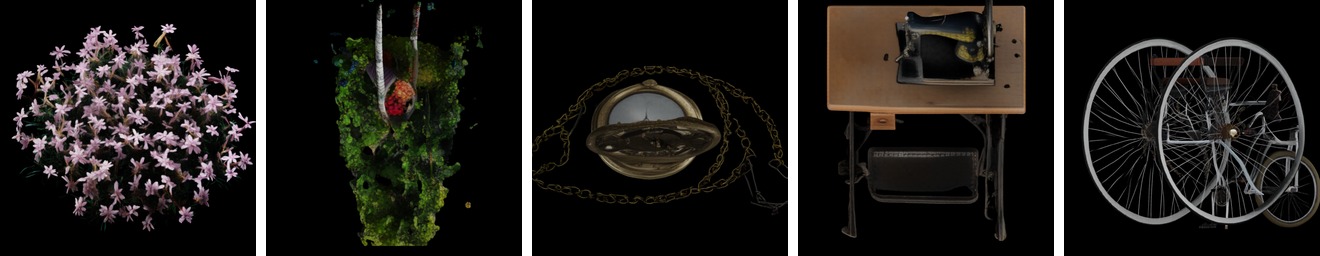} \\
    \parbox[b][\dimexpr0.1629\textwidth\relax][c]{0.13\textwidth}{\raggedright\scriptsize LGM (4 views)} & \includegraphics[width=\linewidth]{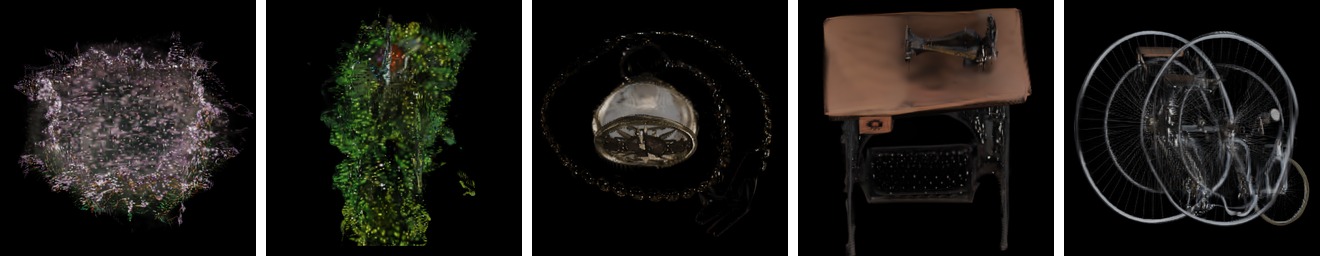} \\
    \parbox[b][\dimexpr0.1629\textwidth\relax][c]{0.13\textwidth}{\raggedright\scriptsize Hunyuan3D-2mv} & \includegraphics[width=\linewidth]{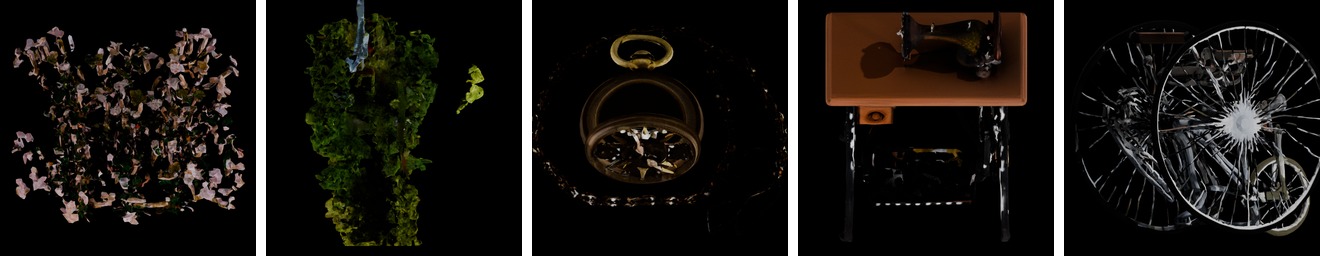} \\
    \parbox[b][\dimexpr0.1629\textwidth\relax][c]{0.13\textwidth}{\raggedright\scriptsize TRELLIS.2-4B} & \includegraphics[width=\linewidth]{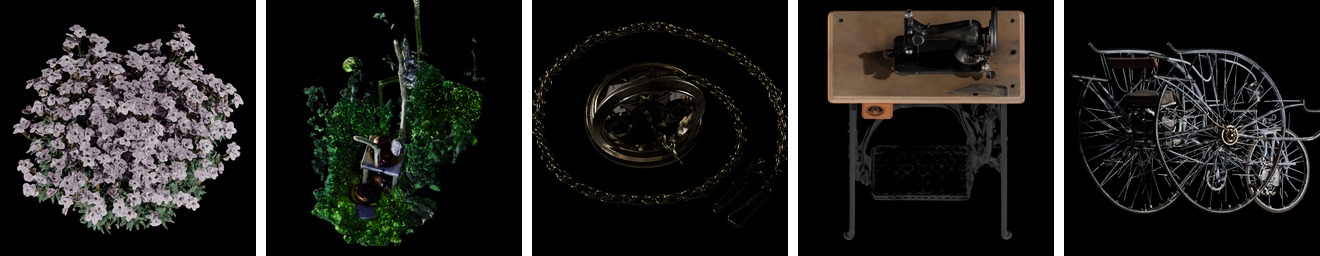} \\
    \parbox[b][\dimexpr0.1629\textwidth\relax][c]{0.13\textwidth}{\raggedright\scriptsize LSRM} & \includegraphics[width=\linewidth]{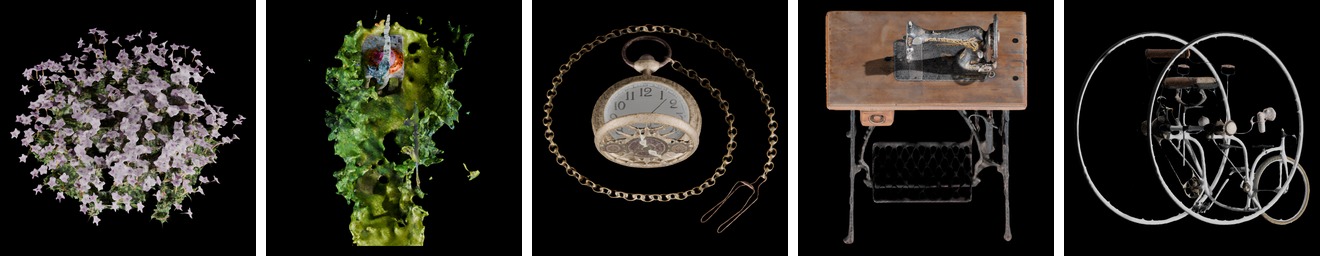} \\
    \parbox[b][\dimexpr0.1629\textwidth\relax][c]{0.13\textwidth}{\raggedright\scriptsize ReLi3D} & \includegraphics[width=\linewidth]{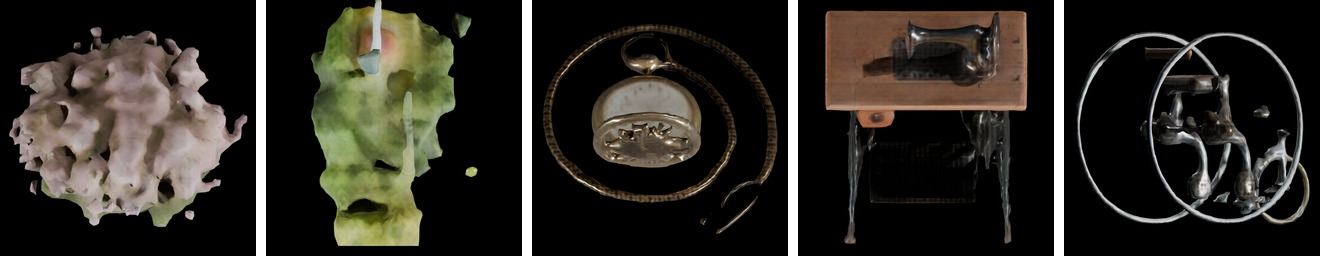} \\
    \parbox[b][\dimexpr0.1629\textwidth\relax][c]{0.13\textwidth}{\raggedright\scriptsize Pixal3D} & \includegraphics[width=\linewidth]{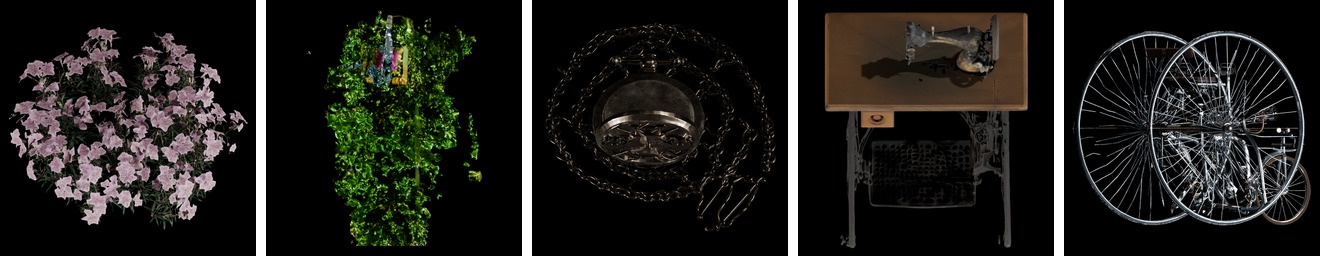} \\
  \end{tabular}
  }
  \caption{Table~\ref{tab:gtcmp}, reconstruction from ground truth views: new views under the new illumination. Columns are objects; rows are ground truth and the methods; within a column every cell shows the same crop of the same view.}
  \label{fig:gtcmp_a_hv}
\end{figure*}

\begin{figure*}[tp]
  \centering\setlength{\tabcolsep}{0pt}
\parbox[b]{0.1390\linewidth}{\centering\scriptsize Input photo}\hspace{0.0037\linewidth}%
\parbox[b]{0.1390\linewidth}{\centering\scriptsize Ground truth}\hspace{0.0037\linewidth}%
\parbox[b]{0.1390\linewidth}{\centering\scriptsize Ours}\hspace{0.0037\linewidth}%
\parbox[b]{0.1390\linewidth}{\centering\scriptsize TRELLIS.2-4B}\hspace{0.0037\linewidth}%
\parbox[b]{0.1390\linewidth}{\centering\scriptsize TRELLIS image-large}\hspace{0.0037\linewidth}%
\parbox[b]{0.1390\linewidth}{\centering\scriptsize DiffSplat}\hspace{0.0037\linewidth}%
\parbox[b]{0.1390\linewidth}{\centering\scriptsize LGM (ImageDream)}%
  \\[1pt]
  \includegraphics[width=\linewidth]{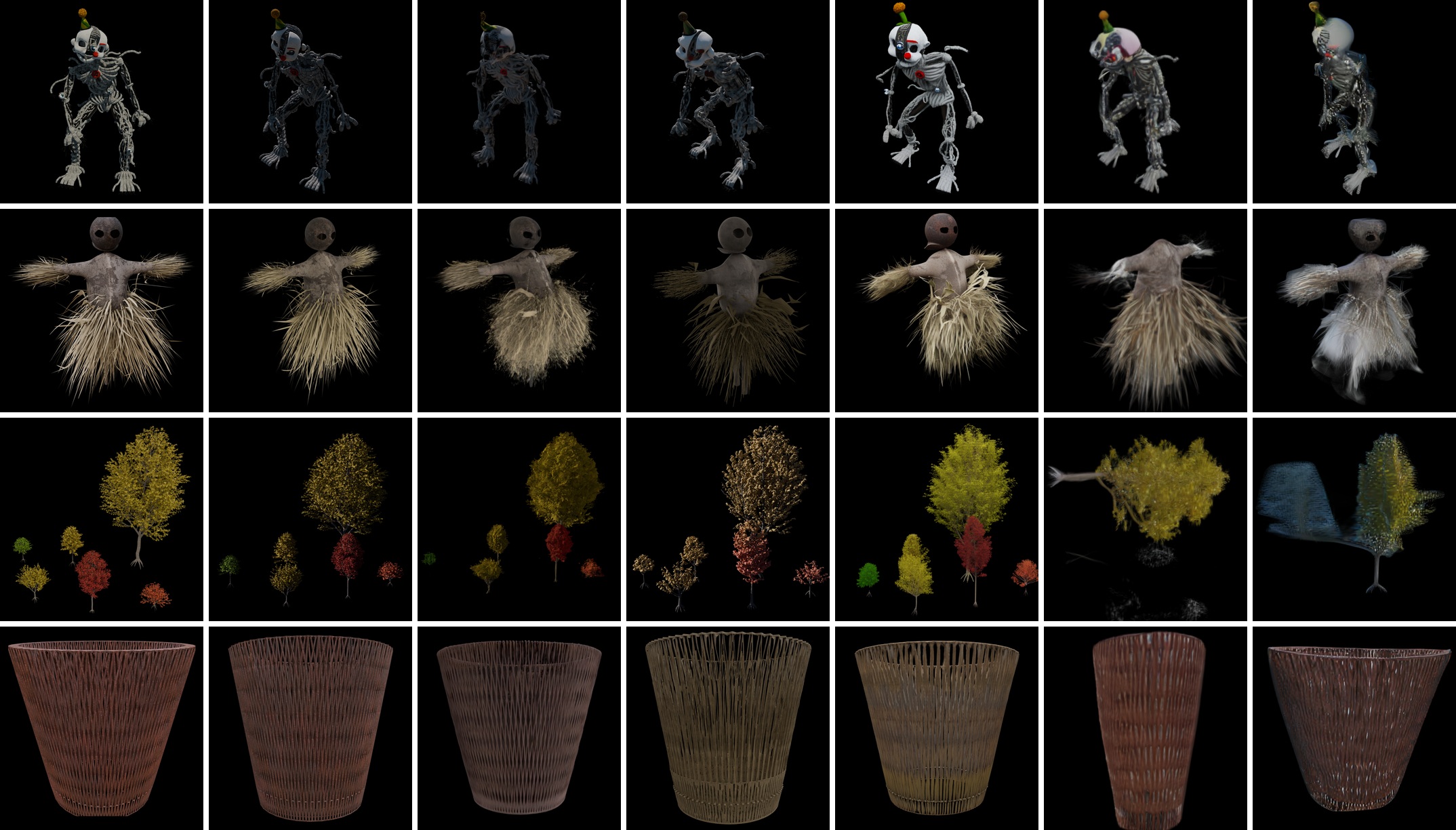}
  \caption{Full single-image pipelines on validation objects under \emph{novel} illumination (one object per row; Table~\ref{tab:gtcmp_single} scores the this table). Left: the input photo, a perspective render of the object under a random environment map. The other cells show a held-out random perspective views, rendered under a different random environment map (random rotation,
intensity) that no method has seen: ground truth, ours and four public single-image generators given the same photo,
rendered with the same camera. }
  \label{fig:gtcmp_single}
\end{figure*}
\begin{table*}[tp]
  \centering\footnotesize\setlength{\tabcolsep}{4pt}
  \caption{Metrics for the objects of Figure~\ref{fig:gtcmp_single} : its 81-view ground truth orbit under the input illumination (left) and the
eight random views under novel environment maps (right). Best values are bold.}
  \label{tab:gtcmp_single}
  \begin{tabular}{@{}l ccc ccc@{}}
    \toprule
     & \multicolumn{3}{c}{Input illumination} & \multicolumn{3}{c}{New illumination} \\
    \cmidrule(lr){2-4}\cmidrule(lr){5-7}
    Method & PSNR$\uparrow$ & SSIM$\uparrow$ & LPIPS$\downarrow$ & PSNR$\uparrow$ & SSIM$\uparrow$ & LPIPS$\downarrow$ \\
    \midrule
    Ours & \textbf{22.15} & \textbf{0.785} & \textbf{0.168} & \textbf{22.72} & \textbf{0.784} & \textbf{0.173} \\
    TRELLIS.2-4B & 19.21 & 0.744 & 0.196 & 19.66 & 0.746 & 0.204 \\
    TRELLIS image-large & 18.38 & 0.752 & 0.195 & 17.45 & 0.740 & 0.202 \\
    DiffSplat (PixArt-$\Sigma$, image-cond.) & 19.10 & 0.759 & 0.201 & 19.87 & 0.757 & 0.209 \\
    LGM (ImageDream) & 19.86 & 0.725 & 0.232 & 19.51 & 0.731 & 0.231 \\
    \bottomrule
  \end{tabular}
\end{table*}

\begin{figure*}[tp]
  \centering\setlength{\tabcolsep}{1pt}\scriptsize
  \newcommand{\rlw}{0.114\linewidth}
  \newcommand{\rlc}[1]{\includegraphics[width=\rlw]{figs/relight/#1.jpg}}
  \begin{tabular}{c@{\hspace{2pt}}cc@{\hspace{4pt}}cc@{\hspace{4pt}}cc@{\hspace{4pt}}cc}
     & \multicolumn{2}{c}{\emph{novel view, training light}} & \multicolumn{2}{c}{\emph{light rotated $90^\circ$}} & \multicolumn{2}{c}{\emph{light rotated $180^\circ$}} & \multicolumn{2}{c}{\emph{light rotated $270^\circ$}} \\
    \cmidrule(lr){2-3}\cmidrule(lr){4-5}\cmidrule(lr){6-7}\cmidrule(lr){8-9}
     & GT & Ours & GT & Ours & GT & Ours & GT & Ours \\
    \rotatebox{90}{\scriptsize ring} & \rlc{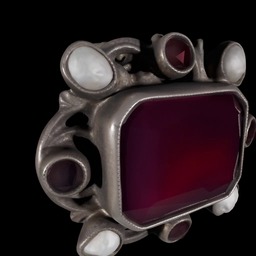} & \rlc{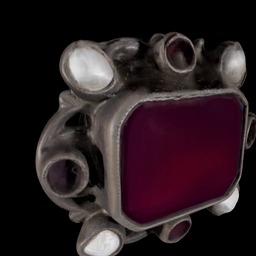} & \rlc{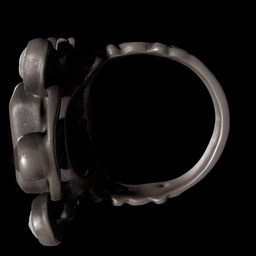} & \rlc{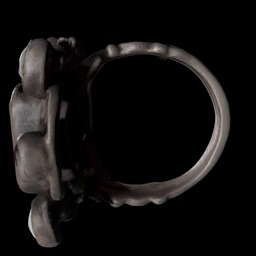} & \rlc{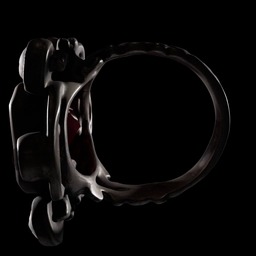} & \rlc{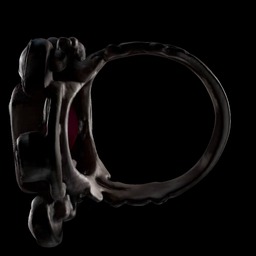} & \rlc{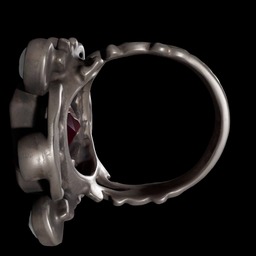} & \rlc{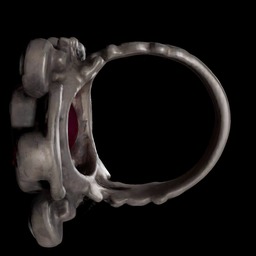} \\
    \rotatebox{90}{\scriptsize chest} & \rlc{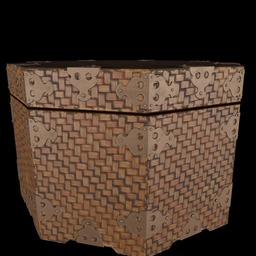} & \rlc{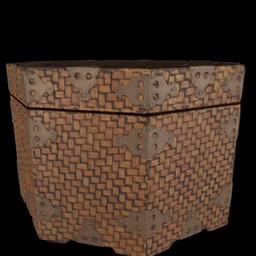} & \rlc{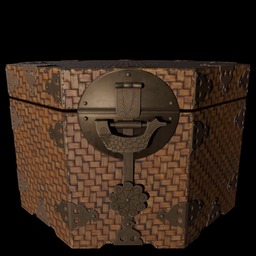} & \rlc{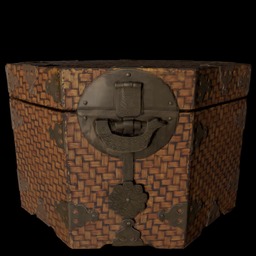} & \rlc{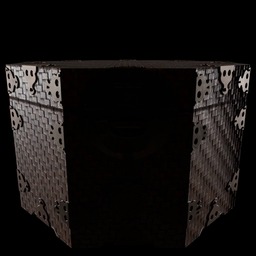} & \rlc{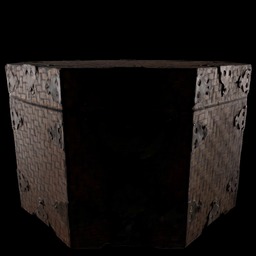} & \rlc{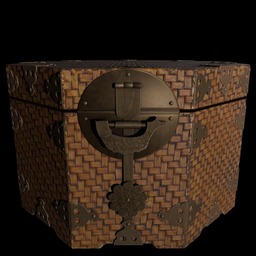} & \rlc{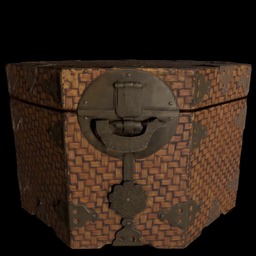} \\
  \end{tabular}
  \caption{View and illumination generalization against Blender ground truth. The reconstructor receives the input turntable. Evaluation uses a $15^\circ$-elevated orbit under the training illumination and a fixed novel view at the same elevation while the training environment map rotates through $360^\circ$. PSNR is averaged over 81 frames per asset, then over 20 assets. All frames shown are novel views: the leftmost pair uses the training environment map, and the remaining pairs rotate that map by $90^\circ$, $180^\circ$, and $270^\circ$. Highlights and shadowed regions follow the ground truth.}
  \label{fig:relight}
\end{figure*}

\paragraph{View and illumination generalization.} Figure~\ref{fig:relight} shows the test described in Section~\ref{sec:exp_main}: a $15^\circ$-elevated novel orbit under the training illumination and a fixed novel view under rotations of the training environment map, compared with Blender ground truth on 20 validation assets. Please note that we train our LRM on a disjoint set of frames from the input frames: the supervision set consists of 64 renderings at fixed and structured trajectories of light and camera across the dataset, as well as 64 per-object random camera and light locations; this increases robustness of the LRM outputs under novel view and illuminations.

\begin{figure*}[tp]
  \centering\setlength{\tabcolsep}{0pt}\scriptsize
  \newlength{\rlw}\setlength{\rlw}{0.112\linewidth}
  \newcommand{\rlHead}[1]{\makebox[\rlw]{#1}}
  \newcommand{\rlRow}[1]{\hbox to1.2em{\hfil\vbox to\rlw{\vfil\hbox{\rotatebox{90}{\scriptsize #1}}\vfil}\hfil}}
  \newcommand{\rlEnv}[2]{\llap{\raisebox{5.05\rlw}[0pt][0pt]{\makebox[\rlw]{\textcolor{white}{#1}}\hspace{#2\rlw}}}}
  \begin{tabular}{@{}c@{\hspace{2pt}}*{4}{c}@{\hspace{8pt}}c@{\hspace{2pt}}*{4}{c}@{}}
     & \multicolumn{2}{c}{Feather} & \multicolumn{2}{c}{Fur coat} & & \multicolumn{2}{c}{Suit mannequin} & \multicolumn{2}{c}{Orca figurine} \\
     & \rlHead{A} & \rlHead{B} & \rlHead{A} & \rlHead{B} & & \rlHead{A} & \rlHead{B} & \rlHead{A} & \rlHead{B} \\
    \vbox{\offinterlineskip\rlRow{Input}\rlRow{\textbf{Ours}}\rlRow{LSRM}\rlRow{Pixal3D}\rlRow{ReLi3D}\rlRow{TRELLIS.2}} &
    \multicolumn{4}{@{}c@{}}{\includegraphics[width=4\rlw]{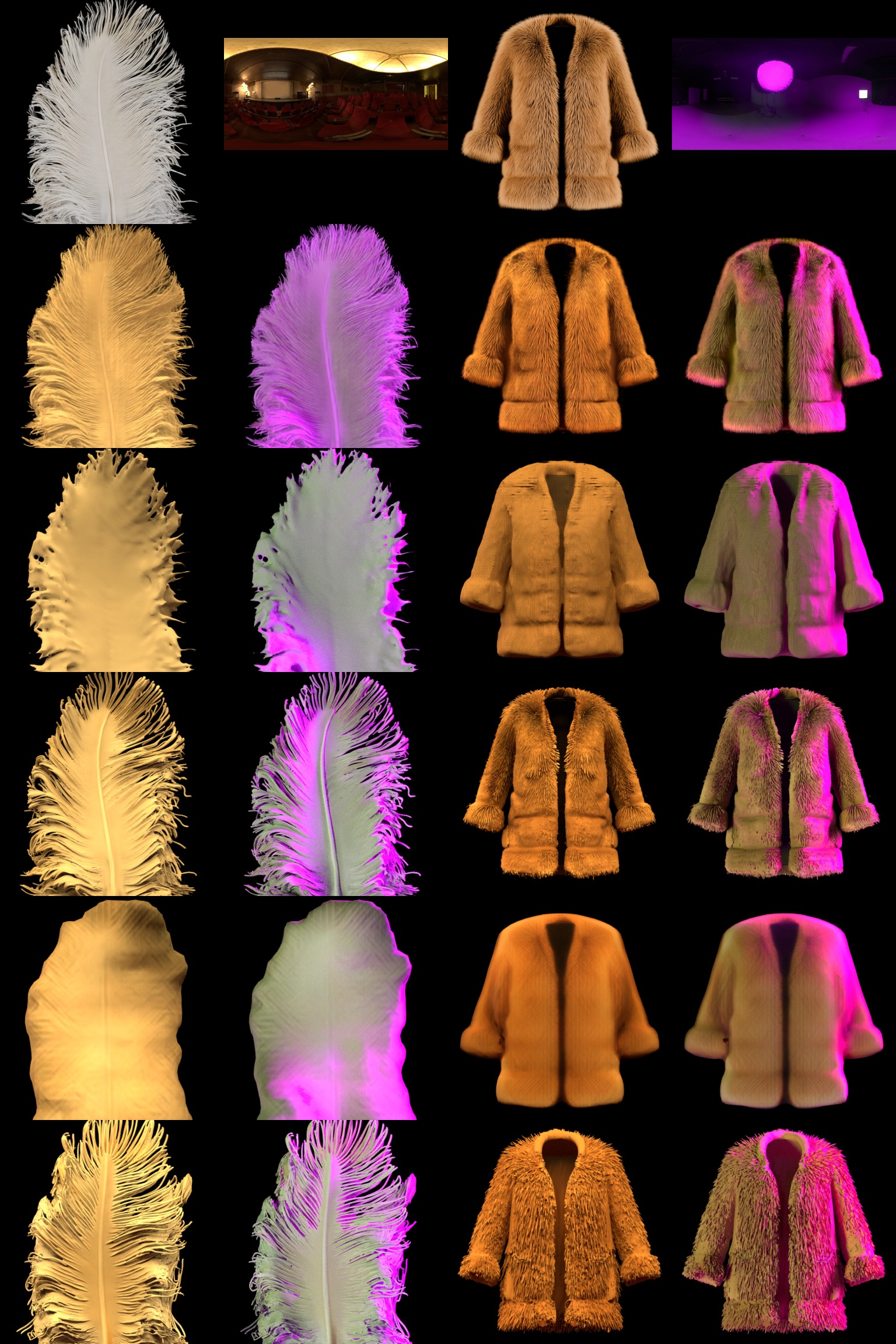}\rlEnv{Environment A}{2}\rlEnv{Environment B}{0}} &
    \vbox{\offinterlineskip\rlRow{Ground truth}\rlRow{\textbf{Ours}}\rlRow{LSRM}\rlRow{Pixal3D}\rlRow{ReLi3D}\rlRow{TRELLIS.2}} &
    \multicolumn{4}{@{}c@{}}{\includegraphics[width=4\rlw]{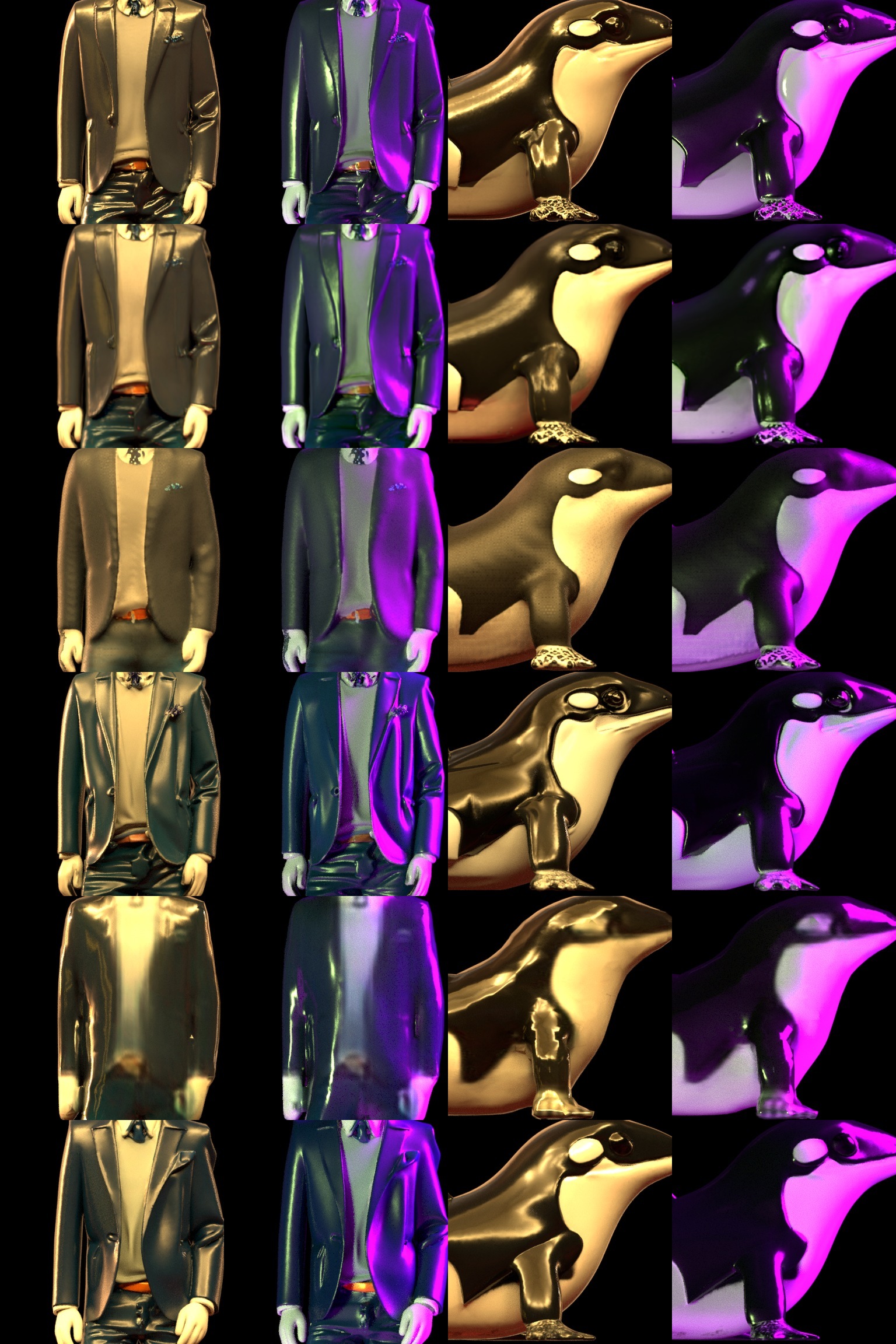}} \\
  \end{tabular}
  \caption{Relighting comparison under two environment maps (A, B). Left: generated objects. Right: objects with ground truth.}
  \label{fig:relight_cmp}
\end{figure*}

\paragraph{Relighting under novel environment maps.} Table~\ref{tab:relight_gt} compares relighting against ground truth with relightable baselines. We use the same 50 complex validation objects as Table~\ref{tab:gtcmp_main}, render each source asset from the front under three environment maps of different color, and render every reconstruction under the same maps. Each image is scaled by one gain per color channel fitted to the ground truth before the metrics are computed. Figure~\ref{fig:relight_cmp} provides separate qualitative comparisons on generated objects and gallery objects with ground truth.

\section{Implementation and Additional Experiments}
\label{sec:implementation}
\subsection{Observation Protocol Details}
\label{sec:observation_details}
The camera is a perspective camera with a $35$\,mm-equivalent focal length (field of view $49.1^\circ$) at distance $1.5$ from the origin of the unit-normalized object, at zero elevation and looking at the origin. Frame $k$ shows the object rotated by $\theta_k=\theta_0-360^\circ\,k/F$, where $\theta_0$ is a fixed calibration offset. We select $29$ uniformly spaced frames, starting at the front view (frame $0$), and encode them into eight latent frames; the first latent holds frame $0$ and each further latent holds four consecutive selected frames, spanning roughly $40^\circ$ of object rotation. One keyframe per latent provides the camera rays that position the emitted Gaussians (Section~\ref{sec:recon}).

\subsection{Training Schedule and Reconstruction Objective}
\label{sec:training_details}
\paragraph{Training.} We train the video generator on approximately 22K assets for three epochs using 128 GPUs. Reconstruction follows a coarse-to-fine schedule: we first fine-tune all DiT parameters for 422.3K steps at $256^2$, followed by an additional 443K steps at $1024^2$. The $1024^2$ stage uses 256 A100 GPUs, with one asset per GPU and one supervision view per asset per step. The reconstructor uses AdamW with a constant learning rate of $2\times10^{-5}$. We apply input noise augmentation with scattered points throughout the region outside the foreground mask.

\paragraph{Losses.} We composite the prediction (using rendered alpha $\hat\alpha$) and reference image (using the dataset's soft alpha $\alpha^*\in[0,1]$) over the same randomly sampled gray background. All image losses are evaluated on these full-frame composites without foreground masking:
\begin{equation}
  \mathcal L_{\mathrm{img}} = 0.8\,\mathcal L_{\ell_2}^{\mathrm{sRGB}}
  + 0.2\,\mathcal L_{\mathrm{LPIPS}}^{\mathrm{VGG}}
  + 0.2\,(1-\mathrm{SSIM}).
  \label{eq:imgloss}
\end{equation}
The silhouette loss is $\mathcal L_{\alpha}=\mathrm{mean}[(\hat\alpha-\alpha^*)^2]$, averaged over all image pixels. Its reference $\alpha^*$ retains soft boundaries, unlike the binary mask used for token selection. The depth loss $\mathcal L_d$ is the $\ell_1$ distance between predicted and ground truth ray depths on foreground pixels. Two weak binarization penalties, $\mathcal L_{\mathrm{bin}}^{\alpha}=\mathrm{mean}[\hat\alpha(1-\hat\alpha)]$ over rendered alpha and $\mathcal L_{\mathrm{bin}}^{o}=\mathrm{mean}_i[\alpha_i(1-\alpha_i)]$ over Gaussian opacities, express a weak preference for values near zero or one; and alpha and opacity remain continuous during optimization. Together they form $\mathcal L_{\mathrm{reg}}$ in Eq.~\ref{eq:total}. The total loss is
\begin{equation}
  \begin{aligned}
  \mathcal L_{\mathrm{total}} &= \mathcal L_{\mathrm{img}}\big(R(d^*,\eta),\,I^*\big) \\
  &+ 1.0\,\mathcal L_{\alpha} + 3.0\,\mathcal L_d(\hat d,\,d^*) \\
  &+ 0.2\,\mathcal L_{\mathrm{bin}}^{\alpha} + 0.01\,\mathcal L_{\mathrm{bin}}^{o}.
  \end{aligned}
  \label{eq:softloss}
\end{equation}
Full-frame compositing and soft-alpha supervision provide training signals for the background and soft boundaries, with the aim of suppressing diffuse Gaussian haze and dark edge halos.

\subsection{Depth-Anchored Appearance Ablation}
\label{sec:exp_depth}

We train three depth-training variants on the full training split at $256^2$: rendering loss only, which jointly optimizes predicted depths and the other Gaussian attributes without a depth loss; joint training with depth supervision (Eq.~\ref{eq:joint}); and depth anchoring (Eq.~\ref{eq:ours}). The two depth-supervised objectives use the same depth-loss term, but supply predicted depths or ground truth depths, respectively, to the training renderer. All models are evaluated using predicted depths, with each asset rendered from a novel viewpoint under a novel lighting condition. Table~\ref{tab:nodepth_full} reports full-frame PSNR, SSIM, and VGG LPIPS; Figure~\ref{fig:nodepth_full} shows two representative examples.

Rendering-loss-only training reaches $24.49$\,dB PSNR, and adding depth supervision raises this to $25.39$\,dB. Depth anchoring reaches $29.22$\,dB, a further gain of $3.83$\,dB over joint training with depth supervision, with better SSIM and LPIPS as well. Figure~\ref{fig:nodepth_full} shows the corresponding reconstructions. This comparison supports using ground truth Gaussian centers as the spatial reference for appearance learning during training, beyond adding depth supervision alone.

\subsection{Shading Rule: Transmittance Blend versus Binary Selection}
\label{app:softblend}

Equation~\ref{eq:shade} blends the two decoder outputs with the traced transmittance $T_{i\ell}$. The shadow-hint formulation of RNA~\citep{rna2023} instead selects one output per light sample by thresholding the visibility at $0.5$, so only the selected output receives gradient. We compare the two rules in a controlled fitting study on 50 assets, fitting each model to these same assets and evaluating full-frame PSNR on validation views of four of them, with everything else fixed: the same data order, initialization, optimizer, and step budget, on 32 GPUs with one sample per GPU. The comparison uses a reduced decoder of four layers and width 64, the size of our distilled per-asset decoders, trained from scratch so that decoder capacity is the bottleneck.

Transmittance blending outperforms binary selection at all seven evaluated checkpoints, reaching $30.62$ versus $29.66$\,dB at 15K steps, a gain of $0.96$\,dB. Two properties explain the direction: the blend is the exact expectation of the direct term under fractional occlusion, and both outputs receive gradient wherever $0<T_{i\ell}<1$, which on a representative asset carries about $21\%$ of the opacity mass, whereas the selection leaves one output without gradient on each side of the threshold.

\subsection{Training: Efficient Foreground Sampling}
\label{sec:efficient}

Foreground sparsity makes the number of tokens and Gaussians a property of the object: across our training set the emitted Gaussian count $N$ per object ranges from about $0.45$\,M (10th percentile) to $8.4$\,M (maximum) at $1024^2$, with a mean of $1.8$\,M, and the cost of a training step---dominated by ray tracing and shading---grows roughly linearly with $N$. With data-parallel training on $G$ GPUs and one object per GPU, every step waits for the largest of $G$ independently drawn objects, so the expected step time is governed by the expected maximum of $N$ rather than its mean; for our training set the expected maximum over 64 draws is $3.3\times$ the mean (Table~\ref{tab:eff}).

We reduce this waiting without changing each object's sampling probability. Objects are sorted by $N$ (available before training from the foreground masks) and split into $K$ equal-size buckets. At each step, all GPUs use a common bucket sampled with probability proportional to its size, and each GPU independently samples an object uniformly within that bucket. The marginal probability of every object is unchanged, so this is still uniform sampling of the training set; only the joint distribution across ranks is altered, so that the $G$ objects processed together have similar cost.

\paragraph{Measured throughput.}
\label{sec:exp_efficiency}

We use $K{=}4$. Table~\ref{tab:eff} quantifies both ingredients. Foreground selection reduces the work per object from $8.4$\,M Gaussians (one per pixel of the eight $1024^2$ views) to $1.8$\,M on average, but with one object per GPU every step waits for the largest of 64 independently drawn objects, whose expectation is $6.1$\,M Gaussians for our size distribution. Bucketed sampling ($K{=}4$) lowers this expected maximum to $3.1$\,M without changing the marginal sampling distribution. On 64 GPUs, it achieves $1.7\times$ the measured training throughput of foreground selection alone. Together with the foreground-only output head, this is what makes training the reconstructor at $1024^2$ with millions of Gaussians per object practical.

\begin{table*}[tp]
  \centering\footnotesize\setlength{\tabcolsep}{4pt}\renewcommand{\arraystretch}{0.95}
  \setlength{\abovecaptionskip}{4pt}\setlength{\belowcaptionskip}{3pt}
  \caption{Training cost at $1024^2$ on 64 GPUs, one asset per GPU. Counts use the 21{,}985 training assets: tokens and per-asset Gaussians are means; per-step Gaussians are the expected maximum over 64 assets. A token covers $16\times16$ pixels. Speedup is relative to foreground selection without bucketing; full-token throughput is unavailable.}
  \label{tab:eff}
  \begin{tabular}{@{}l ccc c@{}}
    \toprule
    Setting & \shortstack{Tokens\\per asset} & \shortstack{Gaussians\\per asset} & \shortstack{Max Gaussians\\per step} & Speedup \\
    \midrule
    Full tokens                                    & 32{,}768 & 8.4\,M & 8.4\,M & -- \\
    Foreground tokens                              & 7{,}937 (24\%) & 1.8\,M & 6.1\,M & $1.0\times$ \\
    Foreground + buckets ($K{=}4$) & 7{,}937 (24\%) & 1.8\,M & 3.1\,M & $1.7\times$ \\
    \bottomrule
  \end{tabular}
\end{table*}

\subsection{Particle Kernel and Asset Conventions}
\label{app:kernel}

The ray tracer evaluates every Gaussian with the generalized Gaussian kernel of 3DGRT~\citep{threedgrt2024}, which Stoch3DGS~\citep{xu2026stoch3dgs} keeps as its default. Along a ray, a particle's response is
\begin{equation}
  \rho(r) \;=\; \exp\!\big(-4.5\,(r/3)^4\big),
  \label{eq:kernel}
\end{equation}
where $r$ is the Mahalanobis distance from the Gaussian center to the ray's closest point, and the particle is ignored where $\rho<e^{-4.5}\approx 0.011$, i.e.\ beyond $3\sigma$. 3DGRT writes the exponent on the squared distance and calls this a degree-2 generalized Gaussian; its implementation refers to the same kernel as degree 4. Compared with the standard Gaussian $\exp(-r^2/2)$, which agrees with Eq.~\ref{eq:kernel} only at the $3\sigma$ cutoff, the kernel is nearly flat within one standard deviation ($0.95$ at $r{=}1$ versus $0.61$) and falls off steeply beyond it. 3DGRT adopted it because denser particles require fewer ray--particle intersections, reporting roughly $2\times$ faster tracing for a loss of about $0.3$\,dB.

\subsection{Appearance Decoder Distillation}
\label{app:decoder_distillation}

\paragraph{Teacher and student.} The full appearance decoder is a nine-layer MLP  with leaky-ReLU of negative slope $0.01$ after every layer, including the six-channel output: lit RGB and shadowed RGB. Global distillation produces a width-128 6 layer decoder shared across assets after 2K steps during training. Separately, per-asset distillation compresses the full decoder into a four-layer student with the same activations for one reconstructed object. The teacher is the decoder associated with that object's reconstruction checkpoint; it is frozen and queried only on the object's own Gaussian features. Gaussian geometry, opacities, and appearance features remain fixed, and only the student MLP is optimized. This compression uses online teacher evaluations without image supervision.

\paragraph{Inputs and sampling.} All decoders in this paper take the same 92-dimensional input. Directions are expressed in each Gaussian's own frame $(\mathbf t,\mathbf b,\mathbf n)$, the columns of $\mathbf R(\mathbf q_i)$ (Section~\ref{sec:rendering}). The input concatenates a Fourier encoding with $L{=}6$ frequencies of the tangent-plane components of $\boldsymbol\omega_{\mathrm{in}}$ and of $\boldsymbol\omega_{\mathrm{out}}$ (26 values each), a Fourier encoding with $L{=}1$ of the tangent-plane components of their half vector $\mathrm{normalize}(\boldsymbol\omega_{\mathrm{in}}+\boldsymbol\omega_{\mathrm{out}})$ (6 values), the raw normal components $\boldsymbol\omega_{\mathrm{in}}\cdot\mathbf n$ and $\boldsymbol\omega_{\mathrm{out}}\cdot\mathbf n$ (2 values), and the 32-dimensional appearance latent, where $\mathrm{Fourier}_L(x)=[\,x,\ \sin(2^k\pi x),\ \cos(2^k\pi x)\,]_{k=0}^{L-1}$ is applied per component. Each distillation step draws $B=65{,}536$ fresh samples: $\boldsymbol\omega_{\mathrm{in}}$ and $\boldsymbol\omega_{\mathrm{out}}$ independently uniform on the sphere, and latents drawn with replacement from the object's Gaussians with probability proportional to opacity.

\paragraph{Distillation objective.} Let $g_\phi$ denote the student and $f_\psi$ the frozen teacher, each producing a six-channel directional response. We minimize their mean absolute difference:
\begin{equation}
  \begin{aligned}
  \mathcal L_{\mathrm{distill}}
  &= \frac{1}{6B}\sum_{b=1}^{B}
  \Big\|g_\phi(\mathbf z_{i_b},\boldsymbol\omega_{\mathrm{in},b},\boldsymbol\omega_{\mathrm{out},b}) \\
  &\quad - f_\psi(\mathbf z_{i_b},\boldsymbol\omega_{\mathrm{in},b},\boldsymbol\omega_{\mathrm{out},b})\Big\|_1.
  \end{aligned}
  \label{eq:decoder_distill}
\end{equation}
We use Adam for 6K steps with a cosine learning-rate schedule from $10^{-3}$ to $10^{-5}$, about $0.39$G teacher evaluations per object. A student is accepted when the full object, shaded once with the teacher and once with the student and composited by the same tracer from two cameras, matches the teacher's rendering within at least $40$\,dB PSNR; typical values are $48$--$49$\,dB. At rendering time, the student replaces the teacher in Eq.~\ref{eq:shade}, retaining both lit and shadowed responses and the same visibility blending.

\begin{figure*}[tp]
  \centering
  \includegraphics[width=\linewidth,height=0.86\textheight,keepaspectratio]{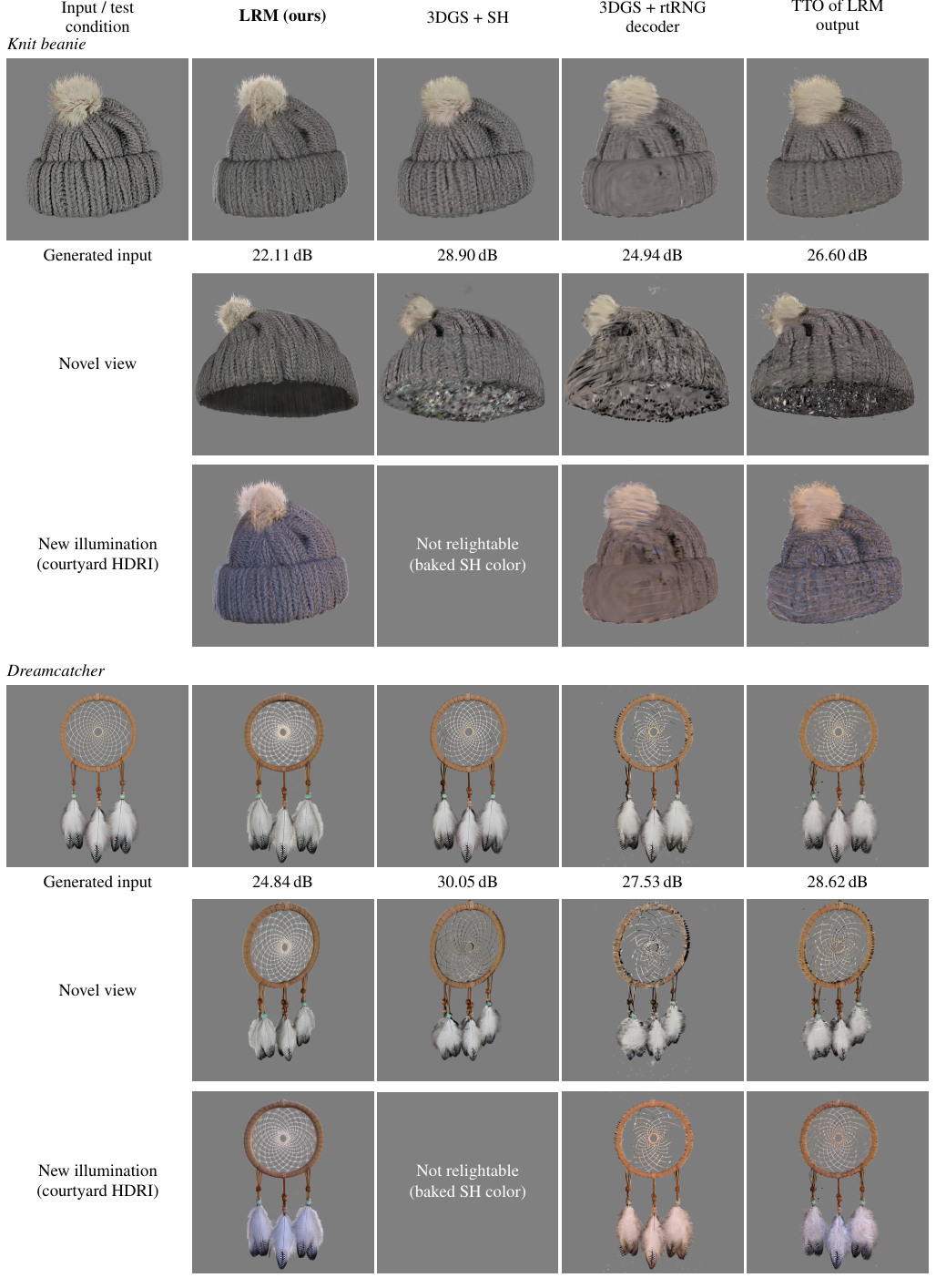}
  \caption{Direct per-object optimization on two generated objects. For each object, rows show an input view, a novel view, and new illumination, with matched cameras and lighting across methods. Numbers report full-frame PSNR over all 81 input frames. The first column shows generated observations, not ground truth. SH appearance cannot be relit.}
  \label{fig:directopt}
\end{figure*}

\subsection{Rendering Performance}
\label{app:rendering_performance}

We benchmark the complete gallery of 62 assets from four views each (248 images), rendered at $1024\times1024$ with 64 samples per pixel on an NVIDIA RTX A6000 GPU. The benchmark uses the OptiX FP32 rendering path, the fourth-order generalized Gaussian kernel with cutoff $3.0$ (Supp.~Sec.~\ref{app:kernel}), point-light appearance decoding, and no cooperative vectors. Table~\ref{tab:rendering_performance} summarizes rendering, asset loading, and geometry acceleration structure (GAS) setup times. Rendering takes $0.872$\,s per view on average, with a 95th-percentile time of $1.436$\,s.

\begin{table}[t]
  \centering\footnotesize\setlength{\tabcolsep}{3pt}\renewcommand{\arraystretch}{0.95}
  \setlength{\abovecaptionskip}{4pt}\setlength{\belowcaptionskip}{3pt}
  \caption{Gallery rendering performance at $1024\times1024$ and 64\,spp on an RTX A6000. All times are in seconds. Rendering statistics are over 248 views; loading and GAS setup statistics are over 62 assets. P95 denotes the 95th percentile.}
  \label{tab:rendering_performance}
  \begin{tabular}{@{}lrrrr@{}}
    \toprule
    Stage & Total & Mean & Median & P95 \\
    \midrule
    Rendering (per view) & 216.26 & 0.872 & 0.798 & 1.436 \\
    Asset loading (per asset) & 112.27 & 1.811 & 1.614 & 3.291 \\
    OptiX/GAS setup (per asset) & 90.04 & 1.452 & 1.422 & 1.860 \\
    \bottomrule
  \end{tabular}
\end{table}

\subsection{Direct Per-Object Optimization}
\label{sec:directopt}
We compare the feed-forward reconstruction with direct per-object optimization against the same generated turntable (81 RGBA frames). Three representations are fitted with the same schedule against all 81 frames at $1024^2$: 3000 Adam iterations, each on a batch of eight frames drawn from a shuffled pass over the sequence (about 300 passes over every frame), and an L1 color term with 0.2\,D-SSIM plus an L1 alpha loss: (i)~3D Gaussians with baked SH color as in 3DGS~\citep{3dgs} (same L1~+~D-SSIM objective and SH degree~3; rasterized with gsplat, densified with MCMC instead of adaptive density control); (ii)~3D Gaussians with an rtRNG relightable decoder of our architecture, trained from scratch together with the Gaussians and rendered with our ray tracer; and (iii)~test-time optimization (TTO) of the LRM's Gaussians and latent features with the shared decoder frozen.

Figure~\ref{fig:directopt} compares a knit beanie and a dreamcatcher. The per-object fits obtain higher input-view PSNR, but this does not ensure coherent appearance away from the fitted observations. At the novel viewpoint, optimized Gaussians show fragmented geometry inside the hat and broken web strands. Under new illumination, the from-scratch decoder and TTO produce smeared knit and feather detail, while the LRM preserves finer structure. SH color bakes in the input illumination and cannot be relit.

\subsection{Limitations}
\label{app:limitations}
The video model sometimes produces inconsistent content across frames, which can lead to ghosting artifacts in reconstructed assets. Accurately reproducing sharp specular effects also remains challenging, as reported in prior neural appearance models~\citep{neumip,rna2023}.

\FloatBarrier
\onecolumn
\section{LLM-Driven Prompt Generation Pipeline}
\label{sec:prompt_construction} We used an AI assistant to generate candidate object descriptions for text-conditioned generation. The task prompt requested independently written descriptions of approximately 40--80 words, each describing a single, plausible real-world object with distinctive overall geometry and inherent geometric or material detail. It encouraged diversity across natural and manufactured subjects, including fibrous, thin, perforated, layered, and optically complex structures, while discouraging template-based expansion and near-duplicate subjects. The instructions excluded camera, lighting, animation, background, and rendering-style directions, since the observation protocol is prescribed by the fine-tuned generator. They also favored structures visible from lateral viewpoints, consistent with the horizontal turntable protocol.

\textbf{Original prompt supplied to the AI assistant.}
The following instructions were used to request candidate descriptions; the requested subject properties do not themselves establish reconstruction capabilities.
\begin{Verbatim}
Create 100 prompts for 3D object generation. Each prompt should describe one interesting, realistic object with distinctive overall geometry and rich, inherent medium-scale and fine-scale geometric or material detail. Prefer subjects with intrinsic optical complexity, such as translucency, subsurface scattering, layered fibers, sheen, iridescence, glossy or wet coatings, or mixed opaque and transmissive regions, as well as other features that are difficult to represent practically with traditional 3D meshes and compact analytic material models.

1. Write each prompt individually.
Independently conceive and write every prompt. Do not use code, scripts, template filling, synonym substitution, or combinatorial expansion to generate or multiply prompt content. Do not pad the collection by repeating the same description with a different name, species, color, or material.

2. Describe only the object itself.
This is because the video model consuming the prompt is finetuned to produce a turntable of only one object. Describe its identity, overall form, proportions, component arrangement, intrinsic coloration, materials, and visible geometric details. Each subject must be one coherent object. A functional assembly with multiple connected parts counts as one object.
Do not describe scenery, backgrounds, ground planes, surrounding props, or added display stands. Integral components, including an object's base, frame, or supports, are allowed if appropriate.
Avoid written text, logos, labels, and legible symbols unless they are essential to the identity or function of the object, since they introduce a separate text-rendering challenge.

3. Exclude all camera, lighting, and animation instructions. This is because the video model consuming the prompt is finetuned to produce a turntable of a static object under pre-defined pose, camera and light conditions, orthogonal to the prompt.
Do not mention cameras, viewpoints, framing, lenses, camera movement, trajectories, turntables, or shot sequences.
Do not describe lighting setups, light sources, cast shadows, moving lights, or changes in lighting and reflections. Intrinsic material properties, such as roughness, metallicity, or translucency, are allowed.
Do not describe actions, behaviors, movement, animation, deformation, or changes over time. Use neutral descriptions of static anatomy, shape, and component arrangement. Distinguish fixed geometry from an action: a helical shell or downward-curving leaf is a shape, not a movement.
A static posture or fixed articulation may be specified when needed to define an animal, plant, vehicle, tool, or mechanical object clearly, but do not describe an action or motion.
When describing optical complexity, state the intrinsic material property itself rather than a lighting setup or how the object changes under moving illumination.
These restrictions are instructions for writing the prompts. Do not repeat them inside the generated prompts, even as negative phrases such as "no camera movement" or "no lighting changes."

4. Choose believable real-world subjects.
Select recognizable natural objects or human-made artifacts. Their anatomy, proportions, materials, thicknesses, connections, and construction must be plausible and internally consistent. Avoid fantastical hybrids, impossible structures, arbitrarily attached parts, and invented embellishments intended solely to increase complexity.

5. Make geometric richness inherent to the subject.
Choose objects whose actual anatomy, material structure, or functional construction already provides interesting geometry. Describe specific features and where they occur. Phrases such as "highly detailed," "intricate texture," are fine but should not be relied upon to create complexity.
Prioritize distinctive medium-scale structures and finer surface relief that can plausibly remain visible at the scale of the complete object. Do not rely on hidden internal mechanisms or invisible microscopic features to justify complexity.
Using more detail than typical for the given object is fine, but do not artificially cover every surface with detail. Preserve naturally smooth regions. Reject objects whose overall geometry is simple and whose apparent complexity comes only from printed patterns, color variation, or added decoration. Simple geometric components are acceptable when their arrangement forms a genuinely rich overall structure, especially when combined with material complexity.
Prefer subjects whose defining structures are visible across lateral viewpoints. Avoid relying on geometry that is visible only from directly above, directly below, or inside the object, since the primary observation protocol is a horizontal turntable.

6. Maximize meaningful diversity across the collection.
Cover a broad range of natural and manufactured subjects, such as terrestrial and marine animals, plants, minerals, foods, textiles, tools, vehicles, instruments, toys, and mechanical devices.
Vary silhouettes, proportions, materials, component arrangements, and types of geometric detail. Related subjects are acceptable when they contribute substantially different geometry. Avoid near-duplicates and excessive concentration in a few categories.
Balance the collection across different representation challenges, including:
- fibrous, hairy, fuzzy, or feathery structures;
- thin, perforated, filamentary, or disconnected geometry;
- branching, layered, folded, frilled, or foliated forms;
- porous, cellular, woven, knitted, braided, or interlocking structures;
- translucent, subsurface-scattering, or mixed opaque and transmissive materials;
- anisotropic, iridescent, pearlescent, glossy, wet, or otherwise strongly view-dependent surfaces; and
- complex but plausible assemblies of multiple materials and connected components.
Individual prompts may exercise several of these challenges. Do not satisfy a category through superficial wording: the relevant structure or material must be inherent to the chosen subject and described concretely.

7. Keep the output concrete and self-contained.
Write each prompt as a concise standalone paragraph of approximately 40-80 words, beginning with a clear object identity. Focus on its defining structures instead of an exhaustive inventory of minor details. Do not add photographic, cinematic, rendering-style, or quality tags.
Before responding, check that each prompt describes only the object, contains no camera or lighting instructions or temporal events, and contributes meaningful structural variety.
Output exactly 100 prompts, one unnumbered prompt per line, with no bullets, category headings, explanations, blank lines, or additional text.
\end{Verbatim}

\end{document}